\documentclass[10pt,twocolumn,letterpaper]{article}

\usepackage[pagenumbers]{cvpr} % To force page numbers, e.g. for an arXiv version

\definecolor{cvprblue}{rgb}{0.21,0.49,0.74}
\usepackage[pagebackref,breaklinks,colorlinks,allcolors=cvprblue]{hyperref}

\def\paperID{*****} % *** Enter the Paper ID here
\def\confName{CVPR}
\def\confYear{2026}

\usepackage{amsmath}
\usepackage{graphicx}
\usepackage{booktabs}
\usepackage{colortbl}
\usepackage{pifont}       % \ding{xx}
\usepackage{bbding}       % \Checkmark,\XSolid,... (需要和pifont宏包共同使用)
\usepackage{fontawesome}  % \faCheck,\faTimes
\usepackage{multirow}
\usepackage{multicol}
\usepackage{tabularx}   % 用于创建宽度可调节的表格
\usepackage{makecell}
\usepackage{float}
\usepackage{amssymb}
\definecolor{Gray}{gray}{0.95}
\definecolor{barriercolor}{RGB}{255, 120, 50}
\definecolor{bicyclecolor}{RGB}{255, 192, 203}
\definecolor{buscolor}{RGB}{255, 255, 0}
\definecolor{carcolor}{RGB}{0, 150, 245}
\definecolor{constructcolor}{RGB}{0, 255, 255}
\definecolor{motorcolor}{RGB}{200, 180, 0}
\definecolor{pedestriancolor}{RGB}{255, 0, 0}
\definecolor{trafficcolor}{RGB}{255, 240, 150}
\definecolor{trailercolor}{RGB}{135, 60, 0}
\definecolor{truckcolor}{RGB}{160, 32, 240}
\definecolor{driveablecolor}{RGB}{255, 0, 255}
\definecolor{otherflatcolor}{RGB}{139, 137, 137}
\definecolor{sidewalkcolor}{RGB}{75, 0, 75}
\definecolor{terraincolor}{RGB}{150, 240, 80}
\definecolor{manmadecolor}{RGB}{213, 213, 213}
\definecolor{vegetationcolor}{RGB}{0, 175, 0}
\definecolor{otherscolor}{RGB}{0, 0, 0}
\definecolor{lightskyblue}{rgb}{0.53, 0.81, 0.98}
\definecolor{aliceblue}{rgb}{0.94, 0.97, 1.0}

\title{LO-World: Unified BEV Latent Modeling for Joint 4D Occupancy and LiDAR
Forecasting}

\author{Zhimin Liao, ~Ping Wei\textsuperscript{\rm}\thanks{Corresponding author.}, ~Kai Mao, ~Shuaijia Chen\\
National Key Laboratory of Human-Machine Hybrid Augmented Intelligence\\
Institute of Artificial Intelligence and Robotics, Xi’an Jiaotong University\\
{\tt\small liaozm@stu.xjtu.edu.cn,~pingwei@xjtu.edu.cn}}

\hypersetup{hypertexnames=false}
\begin{document}
\maketitle
\begin{abstract}
LiDAR point clouds and 3D semantic occupancy represent two distinct modalities capturing fine-grained metric geometry and dense volumetric semantics, respectively, holding natural potential for mutual complementarity. However, existing world models typically forecast these modalities independently or cascade them heuristically, missing the opportunity for reciprocal cross-modal regularization. To address this, we propose \textbf{LO-World}, a unified 4D world model that forecasts both modalities within a shared BEV latent space. This single-stream formulation enables dense semantics and metric geometry to mutually reinforce each other, while eliminating multi-pipeline redundancy. Specifically, LO-World couples an efficient hybrid range-voxel autoencoder with a flow-matching generator that caches historical context to reduce iterative sampling latency. Extensive experiments demonstrate that LO-World outperforms prior state-of-the-art baselines, improving occupancy forecasting IoU by $10.9\%$ and reducing LiDAR Chamfer Distance by $30.1\%$. Meanwhile, it sustains a real-time throughput exceeding $25$~FPS on a single RTX 4090. Our code is available at \url{https://github.com/lzzzzzm/LO-World}.

\end{abstract}

\section{Introduction}
\label{sec:intro}

\begin{figure}[t]
    \centering
    \includegraphics[width=\linewidth, keepaspectratio]{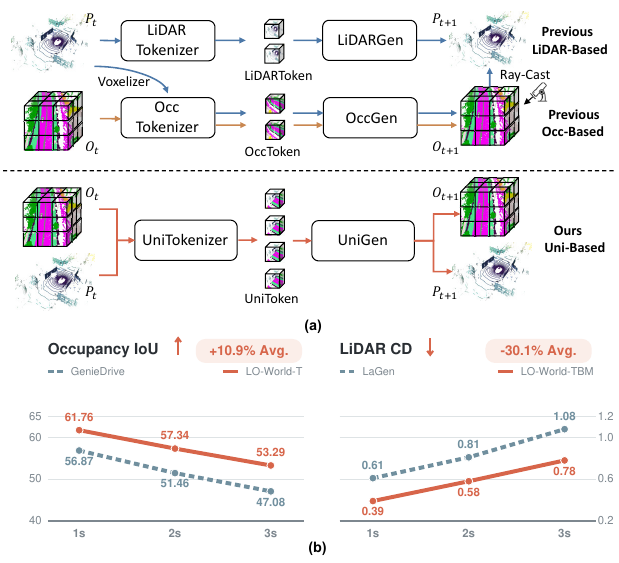}
    \vspace*{-0.2cm}
    \caption{\textbf{(a) Unified forecasting.} LO-World jointly encodes occupancy and LiDAR into a shared BEV latent, forecasting both modalities through a single generative process. 
    \textbf{(b) Performance comparison.} Under comparable control conditions, LO-World achieves a $10.9\%$ relative improvement in 4D IoU and a $30.1\%$ reduction in Chamfer Distance over prior state-of-the-art baselines.}
    \label{figure:introduction_all}
\end{figure}
% !TEX root = ../AnonymousSubmission2027.tex

Autonomous-driving world models~\cite{worldmodel_survey} forecast future 3D scenes to guide downstream tasks~\cite{uniscene,uniscene2,magicdrive,magicdrive-v2, driveworld}. Within these frameworks, 3D semantic occupancy~\cite{occ3d, scene_as_occupancy, occworld, worldliao} and LiDAR point clouds~\cite{semantickitti, waymo, lidarcrafter, copilot4d} offer complementary strengths: occupancy yields dense volumetric semantics but suffers from voxel discretization blur, whereas raw LiDAR captures precise metric geometry despite spatial sparsity. We hypothesize that jointly modeling both modalities enables reciprocal spatio-temporal regularization. Specifically, sharp metric geometry holds the potential to anchor voxel boundaries and alleviate spatial discretization blur, while rich volumetric semantics may guide and stabilize the temporal evolution of sparse point distributions across future frames.

Despite this promising potential, two common paradigms in existing 4D world models~\cite{worldmodel_survey, occworld, worldliao, copilot4d, lidarcrafter} remain limited in exploiting such bidirectional synergy. As illustrated in \cref{figure:introduction_all}a, the first paradigm treats occupancy and LiDAR forecasting as separate tasks, using independent modality-specific pipelines~\cite{occworld, dome, drive-occworld, worldliao, lidarcrafter, copilot4d, VIDAR, R2DM}. When combined to forecast both modalities, these pipelines require separate generative inference and do not explicitly exploit cross-modal structural regularization. The second paradigm, exemplified by UniScene~\cite{uniscene}, adopts an occupancy-first cascade, generating semantic occupancy sequences before synthesizing LiDAR point clouds. However, its stage-wise training prevents downstream LiDAR reconstruction supervision from shaping the upstream occupancy representation, leaving reciprocal geometric and semantic learning underexploited. These limitations raise a pivotal question: \textit{Can we establish an intrinsic connection between dense semantics and fine-grained metric geometry to enable mutual spatio-temporal regularization through a cohesive generative process?}

To answer this question, we present \textbf{LO-World}, a unified generative framework for 4D world modeling. Rather than relying on late-stage attention fusion across isolated branches, LO-World shifts the modeling focus to an intrinsically shared latent space. At the core of our framework is \textbf{LO-VAE}, a multimodal autoencoder~\cite{VAE} designed to bridge the structural gap between the two modalities. Specifically, LO-VAE projects both semantic occupancy and point clouds into a single BEV latent space. Within this pipeline, the LiDAR encoder introduces a hybrid range-voxel architecture to capture native sensor scan patterns before aggregating them into an ego-aligned BEV plane. By embedding metric geometry directly alongside dense semantics, LO-VAE enables deep cross-modal feature synergy, executing joint forecasting through a single generative stream rather than isolated pipelines.

Building upon this shared latent space, we propose \textbf{LO-Former}, an efficient flow-matching generator~\cite{flowmatching} tailored for 4D temporal forecasting. Rather than redundantly encoding historical observations during iterative generation, LO-Former caches the spatio-temporal context once, keeping the repeatedly evaluated flow vector field remarkably lightweight. Coupled with the efficient sampling dynamics of flow matching, this design achieves high-fidelity future rollouts in merely four steps. As shown in \cref{figure:introduction_all}a, LO-World predicts a sequence of unified latents that are decoded into paired semantic occupancy and LiDAR point clouds through a shared decoder with modality-specific heads. Our main contributions are summarized as follows:

\begin{itemize}
\item We explore joint 4D forecasting of 3D semantic occupancy and raw LiDAR point clouds. Empirical results demonstrate that joint modeling can improve forecasting accuracy for both modalities over their unimodal counterparts.

\item We propose LO-World, integrating a hybrid range-voxel autoencoder to forecast both modalities within a single generative stream, alongside a flow-matching generator that reduces iterative sampling overhead through context caching.

\item Extensive evaluations demonstrate the clear superiority of LO-World over prior state-of-the-art baselines. Specifically, it achieves a $10.9\%$ relative gain in occupancy forecasting IoU and a $30.1\%$ error reduction in LiDAR CD, while operating at over $25$~FPS on a single RTX 4090.
\end{itemize}
% !TEX root = ../AnonymousSubmission2027.tex

\section{Related Work}
    \subsection{LiDAR-Based World Models}
        Existing LiDAR world models primarily process point clouds through either range-view projections or 3D voxelization~\cite{4docc,copilot4d,VIDAR,LiDARGen,LiDM,ultralidar,R2DM,lidarcrafter}. Range-view methods~\cite{LiDARGen,LiDM,lagen,R2DM,lidarcrafter}, such as LiDARCrafter~\cite{lidarcrafter} and R2DM~\cite{R2DM}, unfold point clouds onto compact 2D spherical planes to match native beam distributions and alleviate input sparsity. However, pixel proximity on the range image fails to preserve genuine 3D spatial adjacency, introducing geometric distortions. Conversely, voxel-based frameworks~\cite{4docc,copilot4d,VIDAR,ultralidar}, such as Copilot4D~\cite{copilot4d} discretizes 3D space into structured grids to maintain metric neighborhood relations. Nevertheless, faithfully capturing physical structures requires fine-grained voxelization, which incurs prohibitive computational and memory overheads. While multimodal world models such as BEVWorld~\cite{bevworld} explore a unified BEV latent, they rely on standard voxelization for LiDAR and jointly model camera and LiDAR observations. In contrast, LO-World targets the 4D co-forecasting of occupancy and LiDAR through a hybrid range-voxel design that aggregates range-derived features into spatially aligned voxels for subsequent BEV encoding.
    
    \subsection{Occupancy-Based World Models}
        Occupancy provides a dense, semantics-aware representation for 3D world modeling. Following OccWorld~\cite{occworld}, subsequent methods~\cite{renderworld,dfit-occworld,occllama,dome,occsora,worldliao} improve occupancy forecasting through advances in spatial tokenization, temporal modeling, and multi-condition steering. Beyond standalone forecasting, occupancy also serves as an intermediate structural prior for other modalities~\cite{urbandiff,x-scene,4docc,uniscene}. For example, 4DOcc~\cite{4docc} forecasts occupancy from historical LiDAR and renders future point clouds, whereas UniScene~\cite{uniscene} first generates occupancy to condition downstream LiDAR and video generators. Although these pipelines highlight the structural benefit of occupancy, they adopt an occupancy-first formulation rather than jointly forecasting a shared representation of both modalities. In contrast, LO-World jointly encodes spatially aligned occupancy and LiDAR into a shared BEV latent, enabling joint forecasting of both modalities through a single latent-space generative process.

\section{Method}

\begin{figure*}
    \centering
    \includegraphics[
    width=\textwidth,
    height=0.4\textheight,
    keepaspectratio
]{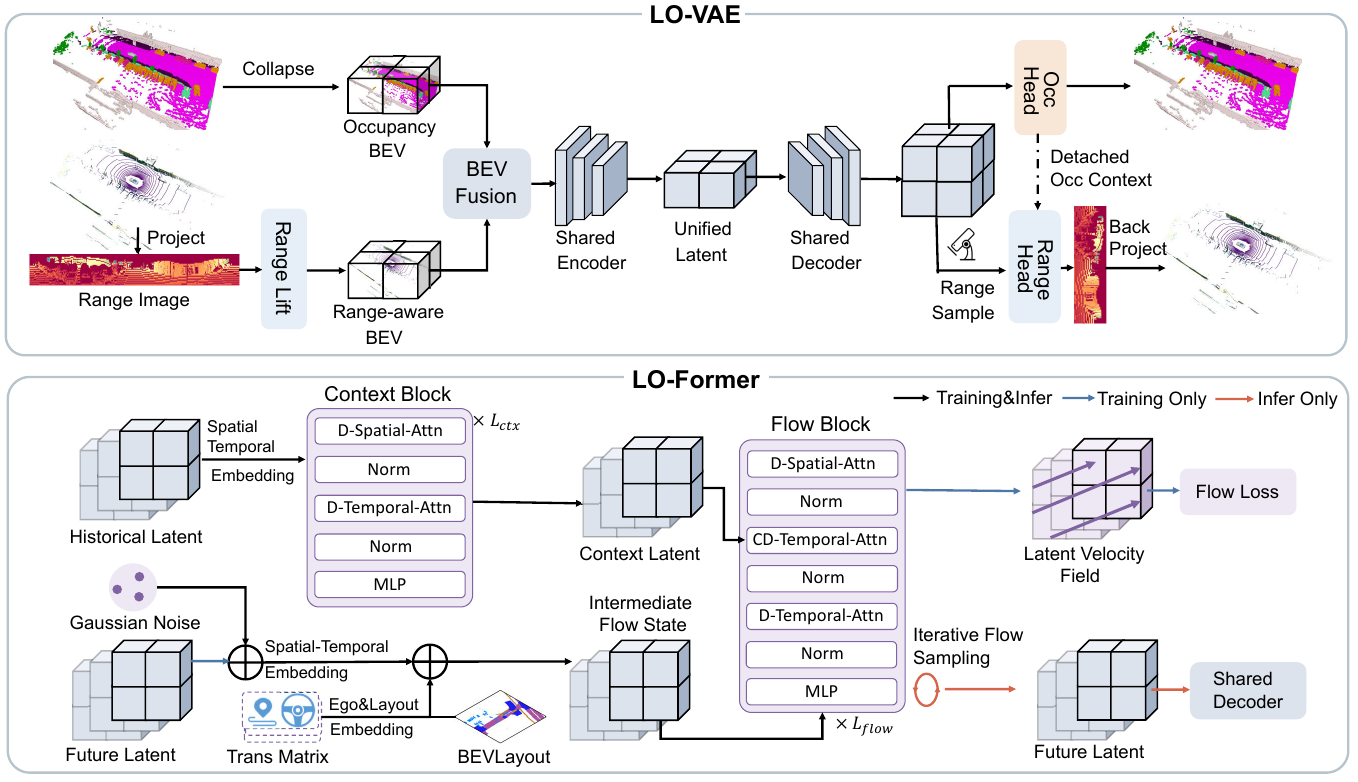}
    \vspace*{-0.1cm}
\caption{\textbf{Overall architecture of LO-World.} Our framework consists of two components: (1) LO-VAE, which jointly tokenizes occupancy and LiDAR into a unified BEV latent space; and (2) LO-Former, a flow-matching-based generation framework that efficiently generates future unified BEV latents from Gaussian noise.}
\label{figure:overall_architecture}
\end{figure*}

\paragraph{Problem Statement.}
We formulate 3D scene evolution as the joint forecasting of semantic occupancy and LiDAR point clouds. 
At each time step $t$, let $\mathcal{P}_{t}$ denote the LiDAR point cloud in the sensor frame and $\mathcal{O}_{t}$ its paired semantic occupancy grid in the ego coordinate system. 
Both modalities are spatially aligned via the known LiDAR-to-ego extrinsic parameters. 
Given $h$ synchronized historical frames, denoted by $\mathbf{P}_{t-h+1:t}=\{\mathcal{P}_{t-h+1},\ldots,\mathcal{P}_{t}\}$ and $\mathbf{O}_{t-h+1:t}=\{\mathcal{O}_{t-h+1},\ldots,\mathcal{O}_{t}\}$, our objective is to forecast both modalities over the subsequent $h'$ time steps. 
For notational brevity, we denote the observation and forecasting horizons as $\tau = \{t-h+1:t\}$ and $\tau' = \{t+1:t+h'\}$, respectively. 
We model this process within a unified predictive framework via the joint conditional distribution:
\begin{equation}
    p(\mathbf{P}_{\tau'}, \mathbf{O}_{\tau'} \mid \mathbf{P}_{\tau}, \mathbf{O}_{\tau}, \mathbf{C}_{\tau'}),
\end{equation}
where $\mathbf{C}_{\tau'}$ denotes the future conditioning signals, which comprise the planned ego-vehicle trajectory alongside optional 3D bounding boxes and map-derived BEV layouts~\cite{magicdrive}.

\paragraph{Overall Architecture.}
As illustrated in~\cref{figure:overall_architecture}, LO-World comprises two key modules: LO-VAE and LO-Former. At each timestamp, LO-VAE aligns and fuses synchronized occupancy and LiDAR features through a shared encoder into a unified BEV latent. Conditioned on past latents and future navigation priors, LO-Former models temporal dynamics via conditional flow matching to generate future latent sequences. Finally, each forecasted latent is decoded through a shared trunk and modality-specific heads to reconstruct both future modalities in a single pass.

\subsection{Unified Tokenization of Occupancy and LiDAR}

\paragraph{Unified BEV Representation.} 

While LiDAR point clouds provide fine-grained metric geometry, 3D semantic occupancy supplies dense, global semantic structure. To harness their mutual complementarity, LO-VAE first establishes a shared BEV coordinate frame that spatially aligns both modalities before encoding them into a unified latent space.

Specifically, let $\mathcal{O}\in\{0,\ldots,K_o-1\}^{H\times W\times Z}$ denote the 3D semantic occupancy grid. Each voxel label is mapped to a learned class embedding. Following prior occupancy encoders~\cite{occworld, worldliao}, we collapse the vertical and embedding dimensions to obtain the occupancy BEV feature $\mathcal{O}_{bev}\in\mathbb{R}^{H\times W\times C_{\mathrm{occ}}}$.

Conventional LiDAR models~\cite{copilot4d,opendwm} typically discretize point clouds directly in 3D Cartesian space, facing an inherent dilemma: coarse voxelization merges adjacent laser beams and destroys native scan patterns, whereas fine grids incur prohibitive computational overhead. To overcome this, we introduce a hybrid range-voxel representation. 

Given a LiDAR scan $\mathcal{P}^{s}$, we first project valid returns onto a 2D spherical range image $\mathcal{R}\in\mathbb{R}^{H_r\times W_r}$ parameterized by azimuth and elevation angles, which naturally preserves native beam topologies and angular resolution. For rays containing multiple returns, the nearest depth is retained. To balance dynamic range disparities while explicitly distinguishing empty regions, we compute a binary validity mask $\mathcal{M}_{p} = \mathbb{I}(\mathcal{R}_{p} > 0)$ and a logarithmically scaled range value:
\begin{equation}
    \hat{\mathcal{R}}_{p} = \frac{\log_{2}(\mathcal{R}_{p} + 1)}{\log_{2}(d_{\max} + 1)}\mathcal{M}_{p},
\end{equation}
where $p$ indexes pixels on $\mathcal{R}$, and $d_{\max}$ denotes the maximum sensor range. Each valid range pixel is thus assigned a two-dimensional feature $\mathbf{f}_{p} = [\hat{\mathcal{R}}_{p}, \mathcal{M}_{p}]^{\top}$.

To spatially align with the occupancy grid, valid range pixels are unprojected into the vehicle ego frame and mapped to corresponding 3D voxels. Each nonempty voxel aggregates its assigned pixel features via average pooling:
\begin{equation}
    \mathbf{V}_{:,k,j,i} = \frac{1}{|\Omega_{k,j,i}|} \sum_{p \in \Omega_{k,j,i}} \mathbf{f}_{p},
\end{equation}
where $\Omega_{k,j,i}$ denotes the set of range pixels falling into voxel coordinates $i$, $j$, and $k$, while unoccupied voxels are padded with zeros. This yields a range-aware feature volume $\mathbf{V}\in\mathbb{R}^{C_v\times Z\times H\times W}$ with $C_v=2$. Finally, collapsing the vertical dimension $Z$ into the channel dimension produces the aligned LiDAR BEV representation $\mathcal{P}_{\mathrm{bev}}\in\mathbb{R}^{H\times W\times C_{\mathrm{lid}}}$, where $C_{\mathrm{lid}}=C_vZ$. Here, voxelization spatially aggregates range-derived features, while subsequent feature encoding operates on the height-collapsed BEV representation, enabling efficient joint encoding with occupancy.

\paragraph{Encoder and Latent Fusion.}
Modality-specific convolutional stems first project $\mathcal{O}_{bev}$ and $\mathcal{P}_{bev}$ to a matched feature dimension. These features are concatenated and fed into a shared ResNet-style encoder~\cite{worldliao}, enabling joint cross-modal interaction across all encoding layers. A variational bottleneck with standard reparameterization~\cite{VAE} subsequently yields the shared BEV latent $\mathbf{z}$.

\paragraph{Unified Decoder and Training Objective.}
LO-VAE upsamples the latent $\mathbf{z}$ via a shared BEV decoder before splitting into modality-specific reconstruction heads. The occupancy head factorizes 3D prediction into binary occupancy classification and multi-class semantic classification, producing semantic logits $\hat{\mathcal{O}}$. To incorporate occupancy structure into LiDAR reconstruction, we condition the LiDAR head on the predicted occupancy probabilities through a stop-gradient connection.

The LiDAR head lifts shared BEV features into 3D ray space. Along each calibrated sensor ray, features sampled from this volume alongside the detached occupancy context are evaluated at $K$ candidate depths to predict depth-bin probabilities and an empty-ray flag. During inference, metric coordinates are directly recovered along rays without dense grid ray-casting.

The training objective unifies occupancy reconstruction, LiDAR range estimation, and variational regularization. The occupancy loss $\mathcal{L}_{occ}$ incorporates Focal Loss~\cite{fbocc} and Lovasz-Softmax Loss~\cite{lovaszloss}:
\begin{equation}
    \mathcal{L}_{occ} = \mathcal{L}_{focal}(\hat{\mathcal{O}}, \mathcal{O}) + \mathcal{L}_{lovasz}(\hat{\mathcal{O}}, \mathcal{O}).
\end{equation}

The LiDAR range loss combines bin cross-entropy with Smooth L1 regression:
\begin{equation}
    \begin{aligned}
    \mathcal{L}_{\mathrm{range}} &= \frac{1}{|\mathcal{R}|} \sum_{p \in \mathcal{R}} w_{p} \mathcal{L}_{\mathrm{CE}}(\hat{\mathbf{y}}_{p}, y_{p}) \\
    &+ \frac{\lambda_{\mathrm{depth}}}{|\mathcal{M}|} \sum_{p \in \mathcal{M}} e_{p} \mathcal{L}_{\mathrm{SL1}}(\tilde{u}_{p}, u(\mathcal{R}_{p})),
    \end{aligned}
    \label{eq:range_loss}
\end{equation}
where $\hat{\mathbf{y}}_{p}$ and $y_{p}$ denote the predicted and ground-truth range-bin labels, $\tilde{u}_{p}$ is the expected normalized log-depth, $w_{p}$ balances ray validity, and $e_{p}$ provides depth-edge weighting. The composite loss is:
\begin{equation}
    \mathcal{L}_{\mathrm{LO\text{-}VAE}} = \mathcal{L}_{\mathrm{occ}} + \lambda_{\mathrm{range}}\mathcal{L}_{\mathrm{range}} + \lambda_{\mathrm{KL}}\mathcal{L}_{\mathrm{KL}},
\end{equation}
where $\lambda_{\mathrm{range}}$ and $\lambda_{\mathrm{KL}}$ balance the respective loss components.

\subsection{Joint Generation of Occupancy and LiDAR}

Conditioned on the historical shared BEV latents $\mathbf{z}_{\tau}$, LO-Former forecasts the future latent sequence $\mathbf{z}_{\tau'}$ via conditional flow matching~\cite{flowmatching}, guided by the planned ego-trajectory alongside optional 3D bounding boxes and HD map layouts~\cite{magicdrive}. 

\begin{table*}[t]
    
    \setlength{\tabcolsep}{0.008\linewidth}
    % \vspace{-4mm}
    \centering
    \scalebox{1.0}{
    \begin{tabular}{l|c|ccc>{\columncolor{Gray}}c|ccc>{\columncolor{Gray}}c|c}
        \toprule
        \multirow{2}{*}{Method} & \multirow{2}{*}{Input/Condition} & \multicolumn{4}{c|}{mIoU (\%) $\uparrow$} & \multicolumn{4}{c|}{IoU (\%) $\uparrow$} & \multirow{2}{*}{FPS $\uparrow$} \\
        & & 1s & 2s & 3s & Avg. & 1s & 2s & 3s & Avg. & \\
        \midrule
        OccWorld-O~\cite{occworld}     & O /Traj.              & 25.78 & 15.14 & 10.51 & 17.14 & 34.63 & 25.07 & 20.18 & 26.63 & 18.00 \\
        % OccLLaMA-O~\cite{occllama}     & O\&Text /Traj.        & 25.05 & 19.49 & 15.26 & 19.93 & 34.56 & 28.53 & 24.41 & 29.17 & - \\
        % DFIT-OccWorld~\cite{dfit-occworld}  & O\&C /Traj.    & 31.68 & 21.29 & 15.19 & 22.71 & 40.28 & 31.24 & 25.29 & 32.27 & - \\
        % DOME~\cite{dome}               & O/Traj.               & 35.11 & 25.89 & 20.29 & 27.10 & 43.99 & 35.36 & 29.74 & 36.36 & 6.54 \\
        UniScene~\cite{uniscene}       & O/Traj.\&Box\&Map     & 35.37 & 29.59 & 25.08 & 30.01 & 38.34 & 32.70 & 29.09 & 33.38 & - \\
        $I^{2}$-World~\cite{worldliao} & O/Traj.               & 47.62 & 38.58 & 32.98 & 39.73 & 54.29 & 49.43 & 45.69 & 49.80 & 37.04 \\   
        GenieDrive~\cite{geniedrive}   & O/Traj.               & 50.47 & 41.47 & 35.83 & 42.59 & 56.87 & 51.46 & 47.08 & 51.80 & 41.38 \\   
        \rowcolor{aliceblue}
        LO-World-T$^{*}$ (Ours)              & O/Traj.            & 52.92 & 44.54 & 38.37 & 45.28 & 57.64 & 53.10 & 49.67 & 53.47 & \textbf{48.25} \\
        \rowcolor{aliceblue}
        LO-World-T (Ours)              & O\&L/Traj.            & 53.43 & 45.21 & 39.39 & 46.01 & 61.76 & 57.34 & 53.29 & 57.46 & 45.20 \\
        \rowcolor{aliceblue}
        LO-World-TBM (Ours)              & O\&L/Traj.\&Box\&Map  & \textbf{59.50} & \textbf{54.78} & \textbf{51.16} & \textbf{55.15} & \textbf{62.52} & \textbf{58.76} & \textbf{55.19} & \textbf{58.82} & 42.46 \\
        \midrule
        % OccWorld-STC                   & C/Traj.                & 10.97 & 15.16 & 10.19 & 7.53  & 20.68 & 24.81 & 20.09 & 17.14 & 3.77 \\
        DOME-STC                       & C/Traj.                & 17.79 & 14.23 & 11.58 & 14.53 & 26.39 & 23.20 & 20.42 & 23.34 & 2.75 \\
        $I^{2}$-World-STC              & C/Traj.                & 21.67 & 18.78 & 16.47 & 18.97 & 30.55 & 28.76 & 26.99 & 28.77 & 4.21 \\
        \rowcolor{aliceblue}
        LO-World-STC (Ours)            & C\&L/Traj.       & \textbf{23.01} & \textbf{20.57} & \textbf{18.46} & \textbf{20.68} & \textbf{32.97} & \textbf{31.77} & \textbf{30.32} & \textbf{31.69} & \textbf{5.01} \\
        \bottomrule
    \end{tabular}
    }
    \caption{\textbf{4D occupancy forecasting performance on the Occ3D-nus validation set.} ``C'', ``O'', and ``L'' denote camera, occupancy, and LiDAR modalities, respectively. $^*$ denotes the single-modality variant of our method.}
    \label{tab:mainresults_occ_forecast}
    % \vspace{-4mm}
\end{table*}

\paragraph{Asymmetric Flow Architecture.}
Standard diffusion and flow-matching models~\cite{ddpm, flowmatching, dit} suffer from considerable inference latency due to the repeated evaluation of deep backbones across iterative sampling steps. To resolve this computational bottleneck, LO-Former adopts a decoupled architecture that separates historical context encoding from recurrent generation via two specialized sub-networks: a deep context encoder and a lightweight flow network. Specifically, the $L_{\mathrm{ctx}}$-layer context encoder processes historical latents in a single forward pass, producing a cached conditional representation that is computed only once per rollout. Conversely, the $L_{\mathrm{flow}}$-layer flow network serves as an iterative generative engine, repeatedly executed at each sampling step to refine future latent trajectories. By amortizing the computational cost of extracting complex spatiotemporal history onto the single-pass context module, the recurrent flow network can be kept remarkably shallow, with $L_{\mathrm{flow}} = 4$ compared to $L_{\mathrm{ctx}} = 8$. This design substantially curtails per-step sampling latency while preserving high generative fidelity.

\paragraph{Context Block and Flow Block.}

The historical latents $\mathbf{z}_{\tau}$ are patchified and augmented with spatial and temporal embeddings. The context block applies deformable spatial attention (DSA)~\cite{deformabel-attn} and deformable temporal attention (DTA)~\cite{bevformer} to aggregate spatiotemporal historical cues into a compact context $\mathbf{H}$. 

Within the flow block, we propose Cross-frame Deformable Temporal Attention (CDTA) as a dedicated bridge connecting cached context to future flow updates. Rather than computing costly dense cross-attention, CDTA uses future tokens as queries to sparsely sample historical features from $\mathbf{H}$ across time. Together with spatial DSA within each future frame and temporal DTA across future frames, CDTA effectively conditions the iterative flow on past dynamics with minimal computational overhead.

\paragraph{Conditional Flow Matching.}

During training, let $\mathbf{x}_{0} = \mathbf{z}_{\tau'}$ denote target future latents and $\boldsymbol{\epsilon} \sim \mathcal{N}(\mathbf{0}, \mathbf{I})$ be standard Gaussian noise. We construct the noisy latent $\mathbf{x}_{\sigma}$ along a linear probability path:
\begin{equation}
    \mathbf{x}_{\sigma} = (1 - \sigma)\mathbf{x}_{0} + \sigma\boldsymbol{\epsilon}, \quad \sigma \in [0, 1].
\end{equation}

Future tokens derived from $\mathbf{x}_{\sigma}$ are augmented with time embeddings, future ego-motion, and BEV layout features~\cite{magicdrive}. The flow network $\hat{\mathbf{v}}_{\psi}$ is trained to predict the constant-velocity field target $\mathbf{v} = \frac{\mathrm{d}\mathbf{x}_{\sigma}}{\mathrm{d}\sigma} = \boldsymbol{\epsilon} - \mathbf{x}_{0}$. The conditional flow-matching objective is defined as:
\begin{equation}
    \mathcal{L}_{\mathrm{FM}} = \mathbb{E}_{\sigma, \mathbf{x}_{0}, \boldsymbol{\epsilon}, \mathbf{c}} \left[ \left\| \hat{\mathbf{v}}_{\psi}(\mathbf{x}_{\sigma}, \sigma, \mathbf{c}) - (\boldsymbol{\epsilon} - \mathbf{x}_{0}) \right\|_{2}^{2} \right],
\end{equation}
where $\mathbf{c}$ includes cached context $\mathbf{H}$, ego-motion, and road layouts.

\paragraph{Efficient Inference.}
During inference, the context $\mathbf{H}$ is pre-computed only once and cached. Starting from standard Gaussian noise $\mathbf{x}_{1} \sim \mathcal{N}(\mathbf{0}, \mathbf{I})$, a deterministic Euler solver integrates the estimated velocity field from $\sigma=1$ to $\sigma=0$. By coupling the lightweight flow block with an efficient four-step sampling schedule, LO-World generates high-fidelity multimodal 4D scenes with remarkable inference speed.

\section{Experiment}
\label{sec:experiments}
\setcounter{dbltopnumber}{1}

\begin{table*}[t]
    \setlength{\tabcolsep}{3pt}
    % \vspace{-4mm}
    \centering
    \scalebox{1.0}{
    \begin{tabular}{l|c|cccc|ccc|c}
        \toprule
        \multirow{2}{*}{Method} & \multirow{2}{*}{Input/Condition} & \multicolumn{4}{c|}{Forecasting Quality($m^{2}$) $\downarrow$} & \multicolumn{3}{c}{Generation Quality} & \multirow{2}{*}{FPS-E2E $\uparrow$} \\
        & & $\mathrm{CD}_{\mathrm{1s}}$ & $\mathrm{CD}_{\mathrm{2s}}$ & $\mathrm{CD}_{\mathrm{3s}}$ & Avg. & FPD $\downarrow$ & B-MMD $\downarrow$ & B-JSD $\downarrow$ \\
        \midrule
        % ViDAR~\cite{VIDAR} & C/Traj. & 1.31 & 1.66 & 2.08 & 1.68 & - & - & - \\
        HERMES~\cite{hermes} & C/Traj. & 0.78 & 0.95 & 1.17 & 0.97 & 105.96 & 21.04 & 4.01 & 0.51 \\
        % 4D-Occ~\cite{4docc} & L/- & 1.19 & 1.48 & 1.85 & 1.50 & - & - & - \\
        OpenDWM-DiT~\cite{opendwm} & L/Traj.\&Box\&Map & 1.96 & 2.78 & 3.21 & 2.65 & 381.91 & 5.73 & 19.90 & 0.10 \\
        LaGen~\cite{lagen} & L/Traj.\&Box\&Map & 0.61 & 0.81 & 1.08 & 0.83 & - & - & 3.97 & - \\
        UniScene~\cite{uniscene} & O/Traj.\&Box\&Map & - & - & - & - & 976.47 & 13.61 & 31.55 & 0.19 \\
        \rowcolor{aliceblue}
        LO-World-T$^{*}$ (Ours) & L/Traj. & 0.46 & 0.81 & 1.18 & 0.82 & 8.84 & 2.68 & 4.54 & \textbf{28.16} \\
        \rowcolor{aliceblue}
        LO-World-T (Ours) & O\&L/Traj. & 0.42 & 0.68 & 0.98 & 0.70 & 5.16 & \textbf{0.76} & 2.96 & 26.26 \\
        \rowcolor{aliceblue} 
        LO-World-TBM$^{*}$ (Ours) & L/Traj.\&Box\&Map & 0.40 & 0.62 & 0.88 & 0.63 & 4.95 & 0.91 & 3.15 & 27.04 \\
        \rowcolor{aliceblue}
        LO-World-TBM (Ours) & O\&L/Traj.\&Box\&Map & \textbf{0.39} & \textbf{0.58} & \textbf{0.78} & \textbf{0.58} & \textbf{4.68} & 0.77 & \textbf{2.94} & 25.31 \\
        % \midrule
        % Oracle & GT & \multicolumn{4}{c|}{-} & \multicolumn{3}{c}{-} \\
        \bottomrule
    \end{tabular}
    }
    \caption{\textbf{4D LiDAR forecasting performance on the nuScenes validation set.} ``C'', ``O'', and ``L'' denote camera, occupancy, and LiDAR, respectively. B-MMD and B-JSD are computed from BEV point-count histograms. Values are multiplied by $10^{4}$ and $10^{2}$, respectively. $^*$ denotes the single-modality variant of our method.}
    \label{tab:mainresults_lidar4d}
    % \vspace{-4mm}
\end{table*}

\begin{table}[tp]
    \setlength{\tabcolsep}{0.013\linewidth}
    \centering
    \scalebox{1.0}{
    \begin{tabular}{l|ccc|cc}
        \toprule
        \multirow{2}{*}{Method}
        & \multicolumn{3}{c|}{Occ. mIoU ($\%$)$\uparrow$}
        & \multicolumn{2}{c}{LiDAR ($m^{2}$)$\downarrow$} \\
        & 4s & 5s & 6s
        & $\mathrm{CD}_{\mathrm{4.0s}}$
        & $\mathrm{CD}_{\mathrm{5.5s}}$ \\
        \midrule
        $I^{2}$-World~\cite{worldliao} & 28.59 & 25.32 & 22.55 & - & - \\
        GenieDrive~\cite{geniedrive} & 31.16 & 27.17 & 23.66 & - & - \\
        LaGen~\cite{lagen} & - & - & - & 1.32 & 1.73 \\
        \rowcolor{aliceblue}
        LO-World-T & 34.01 & 30.70 & 27.48 & 1.49 & 2.12 \\
        \rowcolor{aliceblue}
        LO-World-TBM & \textbf{48.16} & \textbf{45.74} & \textbf{43.52} & \textbf{1.10} & \textbf{1.52} \\
        \bottomrule
    \end{tabular}
    }
    \caption{\textbf{Long-horizon evaluation on nuScenes val set.}}
    \label{tab:mainresults_lh}
\end{table}
\begin{table}[tp]
    \setlength{\tabcolsep}{0.015\linewidth}
    \centering
    \scalebox{1.0}{
    \begin{tabular}{l|cc|cc}
        \toprule
        \multirow{2}{*}{Method} & \multicolumn{2}{c|}{Occupancy ($\%$)$\uparrow$} & \multicolumn{2}{c}{LiDAR ($m^{2}$)$\downarrow$} \\
        & Avg.mIoU & Avg.IoU & $\mathrm{CD}_{\mathrm{1s}}$ & $\mathrm{CD}_{\mathrm{3s}}$ \\
        \midrule
        OpenDWM$^{\dagger}$~\cite{opendwm} & - & - & 2.27 & 3.35 \\
        $I^{2}$-World~\cite{worldliao} & 36.38 & 52.36 & - & - \\
        \rowcolor{aliceblue}
        LO-World-T & 41.31 & 58.30 & 0.91 & 1.77 \\
        \rowcolor{aliceblue}
        LO-World-T$^{\dagger}$ & \textbf{46.00} & \textbf{61.08} & \textbf{0.51} & \textbf{1.28} \\
        \bottomrule
    \end{tabular}
    }
    \caption{\textbf{Performance comparison on the Waymo dataset.} $\dagger$ denotes models fine-tuned on Waymo, while unmarked models are evaluated in a zero-shot setting without target domain fine-tuning.}
    \label{tab:mainresults_zs}
\end{table}

\subsection{Experimental Setup}

\paragraph{Datasets.} Primary evaluations are conducted on nuScenes~\cite{nuscenes} with 32-beam LiDAR and Occ3D-nuScenes~\cite{occ3d} at a $0.4m$ voxel resolution. To evaluate cross-sensor transferability and data-efficient adaptation, we further benchmark on the 64-beam Waymo Open Dataset~\cite{waymo} using Occ3D-Waymo~\cite{occ3d}. Specifically, models are fine-tuned on a tiny split of merely 50 training sequences, representing under $7\%$ of the standard training data, and comprehensively evaluated on the entire 202 validation sequences.

\paragraph{Evaluation Protocol.} Following OccWorld~\cite{occworld}, models observe 2s of history to forecast the subsequent 3s. We benchmark three variants: LO-World-T with trajectories, LO-World-TBM with additional 3D boxes and HD maps, and LO-World-STC incorporating camera-derived STC occupancy~\cite{stcocc}. We report non-empty IoU and semantic mIoU for occupancy forecasting, and CD for predictive LiDAR geometry. We assess the distributional fidelity of forecasted point
clouds using Fr\'echet Pointcloud Distance (FPD)~\cite{fpd},
BEV-based MMD (B-MMD)~\cite{mmd}, and JSD
(B-JSD)~\cite{mmd}. Computational efficiency is reported using model FPS and FPS-E2E.

\paragraph{Implementation Details.} All LO-World variants are trained on eight NVIDIA RTX 4090 GPUs, while the downstream video model uses eight A100 GPUs. Following OccWorld~\cite{occworld}, model FPS is evaluated on a single RTX 4090 with a batch size of one, measuring current-frame encoding and latent forecasting without decoding. FPS-E2E additionally includes dual-modality decoding of all six predicted future frames.

\subsection{Main Results}

We additionally evaluate unimodal-input variants that
use either historical occupancy or historical LiDAR
as the sole observed modality.

\begin{table}[t]
    \centering
    \setlength{\tabcolsep}{2.2pt}
    \renewcommand{\arraystretch}{1.12}
    \resizebox{\columnwidth}{!}{
    \begin{tabular}{@{}l|c|c|ccc@{}}
        \toprule
        Method & Frames & Condition
        & FVD$\downarrow$
        & mIoU$\uparrow$
        & mAP$\uparrow$ \\
        \midrule
        MagicDrive~\cite{magicdrive}
        & 60 & BEV + Box & 217.94 & 18.27 & 11.86 \\
        UniScene~\cite{uniscene} & 32 & 2\,Hz GT Occ.  & 610.15 & 18.59 & 6.24 \\
        GenieDrive~\cite{geniedrive} & 81 & 2\,Hz GT Occ.  & 92.78 & 31.03 & 18.89 \\
        \midrule
        \rowcolor{aliceblue}
        Ours
        & 81 & 12\,Hz Pred. Occ.
        & \textbf{60.41} & \textbf{34.83} & \textbf{27.15} \\
        \midrule
        \rowcolor{gray!10}
        Oracle
        & 81 & 12\,Hz GT Occ. & \textit{40.78} & \textit{36.47} & \textit{30.24} \\
        \bottomrule
    \end{tabular}
    }
    \caption{\textbf{Downstream video generation on the nuScenes validation set.}  
    “Oracle” denotes the ground-truth-conditioned reference using 12 Hz occupancy.}
    \label{tab:downstream}
\end{table}

\paragraph{4D Occupancy Forecasting.} As shown in \cref{tab:mainresults_occ_forecast}, our occupancy-only variant achieves the highest average mIoU and IoU among the evaluated occupancy-only baselines. Incorporating LiDAR input through LO-World-T improves average IoU from $53.47\%$ to $57.46\%$ and mIoU from $45.28\%$ to $46.01\%$. With this additional modality, LO-World-T outperforms GenieDrive by $8.03\%$ in mIoU and $10.93\%$ in IoU in relative terms, while achieving $45.20$ model FPS. Adding box and map priors raises mIoU to $55.15\%$ for LO-World-TBM, a $19.9\%$ relative improvement over LO-World-T. Finally, LO-World-STC, which additionally uses LiDAR input, outperforms the camera-based $I^{2}$-World-STC by $9.0\%$ in mIoU and $10.1\%$ in IoU in relative terms.

\paragraph{4D LiDAR Forecasting.} As shown in \cref{tab:mainresults_lidar4d}, the unimodal LO-World-T$^{*}$ variant achieves an average CD of $0.82$. Joint modeling with semantic occupancy reduces CD to $0.70$ for LO-World-T, demonstrating improved LiDAR forecasting. Incorporating box and map guidance further lowers CD to $0.58$ for LO-World-TBM, a $30.1\%$ relative reduction compared with LaGen~\cite{lagen}. Compared with its LiDAR-only counterpart, LO-World-T
also achieves lower FPD, B-MMD, and B-JSD
($2.96$ vs.\ $4.54$ for B-JSD). Finally, LO-World-T and LO-World-TBM achieve $26.26$ and $25.31$ FPS-E2E, respectively, on a single RTX 4090.

\begin{table}[tp]
    \centering

    % Ablation architecture illustration
    \includegraphics[width=\linewidth]{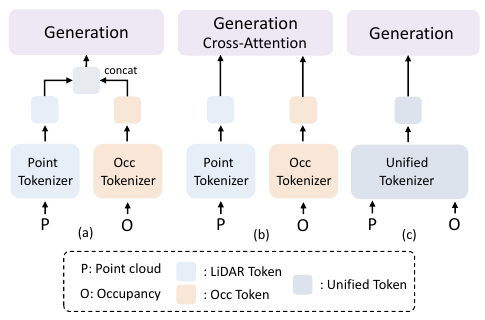}

    \vspace{0.5em}

    % Ablation results
    \setlength{\tabcolsep}{0.003\linewidth}
    \resizebox{\columnwidth}{!}{
    \begin{tabular}{l|c|c|cc|c}
        \toprule
        \multirow{2}{*}{Method}
        & \multirow{2}{*}{\shortstack{Avg.\\IoU ($\%$)$\uparrow$}}
        & \multirow{2}{*}{\shortstack{Avg.\\CD ($m^{2}$)$\downarrow$}}
        & \multirow{2}{*}{FPD $\downarrow$}
        & \multirow{2}{*}{JC $\uparrow$}
        & \multirow{2}{*}{FPS-E2E $\uparrow$} \\
        & & & & & \\
        \midrule
        (a) & 54.08 & 0.84 & 9.01 & 0.52 & 17.76 \\
        (b) & 54.71 & 0.80 & 8.94 & 0.55 & 10.24 \\
        \rowcolor{aliceblue}
        (c) & \textbf{57.46} & \textbf{0.70} & \textbf{5.16} & \textbf{0.77} & 26.26 \\
        \midrule
        w/o Hb & 50.01 & 0.92 & 75.23 & 0.65 & \textbf{26.30} \\
        w/o Hb$^{\diamond}$ & 55.87 & 0.75 & 32.59 & 0.67 & 6.72 \\
        w/o Sem & -- & 0.97 & 56.12 & 0.60 & 26.27 \\
        \bottomrule
    \end{tabular}
    }

    \caption{\textbf{Ablation study on the unified multimodal design.}
    ``Hb'' indicates the hybrid range-voxel representation for LiDAR,
    ``Sem'' specifies whether semantic labels are incorporated into
    the occupancy volume, and $^{\diamond}$ denotes the
    high-resolution voxelization setting.}

    \label{table:ablation_study_unified_vae}
\end{table}
\paragraph{Long-Horizon Forecasting.} As reported in \cref{tab:mainresults_lh}, we further benchmark temporal scalability across $4\text{--}6\,\mathrm{s}$ horizons. In occupancy forecasting, LO-World-T surpasses GenieDrive by $12.4\%$ in mean mIoU, while LO-World-TBM maintains high mIoU scores exceeding $43\%$ through the entire $6\,\mathrm{s}$ rollout. In LiDAR forecasting, LO-World-TBM achieves lower long-horizon CD than LaGen at both evaluated horizons, while the trajectory-only variant remains less accurate in LiDAR forecasting.

\begin{table}[t]
    \centering
    \setlength{\tabcolsep}{4pt}
    \renewcommand{\arraystretch}{1.08}
    \begin{tabular}{l|ccc}
        \toprule
        Method & Avg. mIoU$\uparrow$ & Avg. CD$\downarrow$ & FPS-E2E$\uparrow$ \\
        \midrule
        \multicolumn{4}{c}{\textit{Architecture Ablation}} \\
        \midrule
        \rowcolor{aliceblue}
        Full-Model & \textbf{46.01} & \textbf{0.70} & 26.26 \\
        w/o Context & 37.19 & 1.58 & \textbf{34.34} \\
        w/o Context$^{\ddagger}$ & 43.60 & 0.74 & 18.43 \\
        \midrule
        \multicolumn{4}{c}{\textit{Layers Ablation (Context, Flow)}} \\
        \midrule
        (4, 4) & 42.81 & 0.90 & 31.71 \\
        (6, 4) & 44.52 & 0.78 & 30.44 \\
        \rowcolor{aliceblue}
        (8, 4) & 46.01 & 0.70 & 26.26 \\
        (8, 6) & \textbf{46.36} & \textbf{0.68} & 24.12 \\
        (8, 2) & 40.85 & 0.83 & \textbf{35.53} \\
        \midrule
        \multicolumn{4}{c}{\textit{Sampling Steps}} \\
        \midrule
        1 step & \textbf{47.20} & 0.83 & \textbf{40.06} \\
        2 steps & 47.19 & 0.83 & 35.72 \\
        \rowcolor{aliceblue}
        4 steps & 46.01 & 0.70 & 26.26 \\
        8 steps & 43.80 & \textbf{0.68} & 21.54 \\
        \bottomrule
    \end{tabular}
    \caption{\textbf{Ablation studies on LO-Former.} $^{\ddagger}$ indicates the depth-matched variant with 12 flow blocks and no context blocks.}
    \label{tab:ablation_loformer}
\end{table}

\paragraph{Generalization Ability.} As presented in \cref{tab:mainresults_zs}, zero-shot LO-World-T surpasses $I^{2}$-World by over $11\%$ in both mIoU and IoU, while achieving lower CD than fine-tuned OpenDWM~\cite{opendwm}. Fine-tuning on 50 Waymo sequences further increases mIoU to $46.00\%$ and reduces CD at 1\,s from $0.91$ to $0.51$, supporting cross-dataset transfer and adaptation with limited target-domain training data.

\paragraph{Downstream Video Generation.}
As shown in~\cref{tab:downstream}, we fine-tune the GenieDrive~\cite{geniedrive} video generation framework using the 12\,Hz occupancy predicted by LO-World-TBM. Our approach achieves an FVD of $60.41$ over 81 frames, alongside $34.83\%$ mIoU and $27.15\%$ mAP, improving over the GenieDrive baseline conditioned on 2\,Hz ground-truth occupancy. These results demonstrate the downstream utility of our predicted occupancy sequences, while a gap remains to the reference conditioned on 12\,Hz ground-truth occupancy.

\begin{figure*}[!t]
    \centering
    \includegraphics[width=\textwidth]{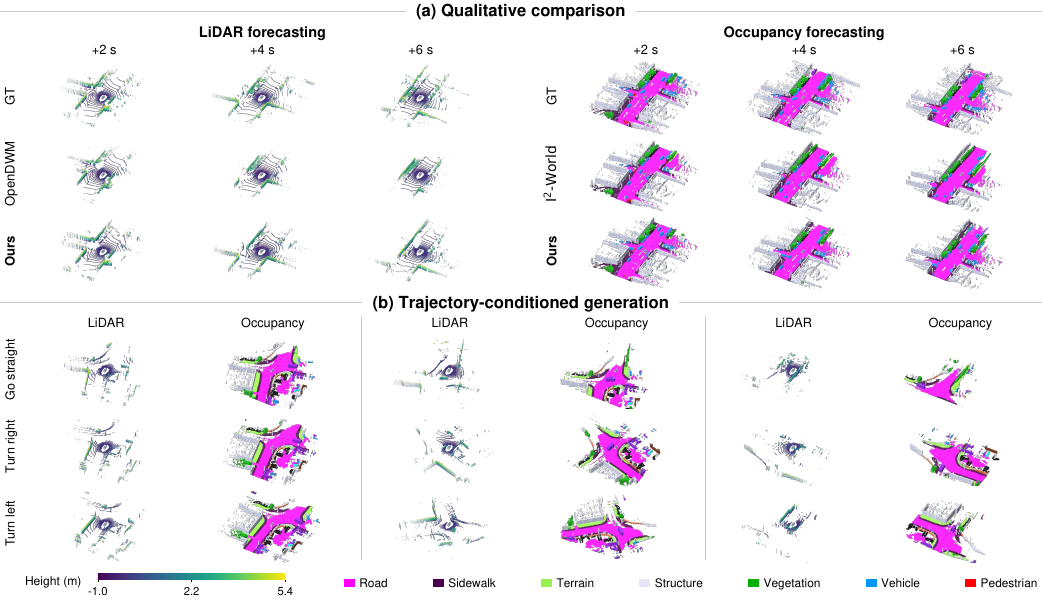}%
    \vspace*{-0.2cm}
\caption{\textbf{Qualitative forecasting on nuScenes.} (a) LiDAR and occupancy forecasts at 2, 4, and 6\,s versus OpenDWM and \mbox{$I^{2}$-World}, respectively. (b) Paired forecasts under straight, right-turn, and left-turn ego-vehicle trajectories.}

\label{figure:qualitative_results}
\end{figure*}

\subsection{Ablation Study}

All ablations use LO-World-T unless stated otherwise.

\paragraph{Ablation on Latent and Representation Designs.}
In \cref{table:ablation_study_unified_vae}, we compare our shared-latent design with two baselines using separate modality-specific latents: (a) employs separate encoders and concatenates modality tokens before forecasting, while (b) introduces cross-modal attention into the context and flow-matching blocks. In contrast, LO-VAE maps both modalities into a shared BEV latent space, yielding higher occupancy IoU and lower LiDAR CD and FPD than both baselines while achieving higher end-to-end throughput: $26.26$ FPS-E2E versus $17.76$ and $10.24$, respectively. To evaluate cross-modal spatial agreement beyond independent metrics, we introduce Joint Consistency (JC), defined as the harmonic mean of surface recall and precision relative to reference geometry (details in supplementary). Benefiting from intrinsic alignment, our shared-latent design achieves a JC of $0.77$, substantially outperforming $0.52$ for baseline~(a) and $0.55$ for baseline~(b).

Furthermore, we assess the hybrid range-voxel LiDAR representation. Replacing it with standard voxelization~\cite{copilot4d, opendwm} increases CD from $0.70$ to $0.92$. Increasing the voxel resolution reduces CD to $0.75$ but lowers throughput from $26.30$ to $6.72$ FPS-E2E. These results support the favorable quality--efficiency trade-off of our hybrid representation. Lastly, removing semantic labels from occupancy increases CD from $0.70$ to $0.97$ and FPD from $5.16$ to $56.12$, supporting the benefits of semantic information for both forecasting accuracy and distributional fidelity.

\paragraph{Ablation on LO-Former Configurations.}
In \cref{tab:ablation_loformer}, we ablate the context--flow architecture, layer allocations, and sampling steps. Removing context encoding reduces mIoU and increases CD, while a depth-matched baseline with 12 iterative flow blocks yields lower accuracy and throughput, supporting the accuracy--efficiency benefits of decoupled context encoding and iterative flow updates. With eight context blocks, increasing the flow depth from four to six blocks slightly improves mIoU and CD but reduces throughput from $26.26$ to $24.12$ FPS-E2E. For sampling steps, increasing from one to four steps reduces CD by $15.7\%$, at the cost of a $2.5\%$ relative decrease in mIoU and lower throughput. Increasing from four to eight steps yields only a further $2.9\%$ reduction in CD, while incurring a $4.8\%$ relative decrease in mIoU and further reducing throughput. We therefore adopt four steps as a balanced default for joint forecasting.

\subsection{Visualizations}
\paragraph{Qualitative Comparison and Controllability.}
As illustrated in \cref{figure:qualitative_results}, LO-World produces more complete object contours and more coherent scene layouts than OpenDWM for LiDAR and $I^{2}$-World for occupancy in the shown examples. Furthermore, conditioning the same historical context on different ego-vehicle trajectories produces distinct future forecasts across both modalities, illustrating the trajectory-conditioned controllability of LO-World.

\section{Conclusion}
In this paper, we present \textbf{LO-World}, a unified 4D world model that forecasts 3D semantic occupancy and LiDAR point clouds within a shared BEV latent space. By integrating a hybrid range-voxel autoencoder with a context-caching flow-matching generator, LO-World realizes high-fidelity rollouts in merely four sampling steps. Experiments demonstrate improved forecasting accuracy for both modalities over the corresponding single-modality variants, with forecasting throughput exceeding $25$ frames per second on a single RTX 4090. This unified formulation offers a promising paradigm for future autonomous-driving world models.

{
    \small
    \bibliographystyle{ieeenat_fullname}
    \bibliography{main}

@String(CVPR= {IEEE Conf. Comput. Vis. Pattern Recog.})

@String(ICCV= {Int. Conf. Comput. Vis.})

@String(ECCV= {Eur. Conf. Comput. Vis.})

@String(NIPS= {Adv. Neural Inform. Process. Syst.})

@String(ICLR = {Int. Conf. Learn. Represent.})

@String(AAAI = {AAAI})

@String(CVPR  = {CVPR})

@String(ICCV  = {ICCV})

@String(ECCV  = {ECCV})

@String(NIPS  = {NeurIPS})

@String(ICLR  = {ICLR})

@inproceedings{uniscene,
  title={Uniscene: Unified occupancy-centric driving scene generation},
  author={Li, Bohan and Guo, Jiazhe and Liu, Hongsi and Zou, Yingshuang and Ding, Yikang and Chen, Xiwu and Zhu, Hu and Tan, Feiyang and Zhang, Chi and Wang, Tiancai and others},
  booktitle={CVPR},
  year={2025}
}

@article{opendwm,
  title={Unimlvg: Unified framework for multi-view long video generation with comprehensive control capabilities for autonomous driving},
  author={Chen, Rui and Wu, Zehuan and Liu, Yichen and Guo, Yuxin and Ni, Jingcheng and Xia, Haifeng and Xia, Siyu},
  journal={arXiv preprint},
  year={2024}
}

@inproceedings{LiDARGen,
  title={Learning to generate realistic lidar point clouds},
  author={Zyrianov, Vlas and Zhu, Xiyue and Wang, Shenlong},
  booktitle={ECCV},
  year={2022},
}

@inproceedings{LiDM,
  title={Towards realistic scene generation with lidar diffusion models},
  author={Ran, Haoxi and Guizilini, Vitor and Wang, Yue},
  booktitle={CVPR},
  year={2024}
}

@inproceedings{R2DM,
  title={Lidar data synthesis with denoising diffusion probabilistic models},
  author={Nakashima, Kazuto and Kurazume, Ryo},
  booktitle={ICRA},
  year={2024},
}

@article{lidarcrafter,
  title={LiDARCrafter: Dynamic 4D world modeling from LiDAR sequences},
  author={Liang, Ao and Liu, Youquan and Yang, Yu and Lu, Dongyue and Li, Linfeng and Kong, Lingdong and Zhao, Huaici and Ooi, Wei Tsang},
  journal={arXiv preprint},
  year={2025}
}

@article{copilot4d,
  title={Copilot4d: Learning unsupervised world models for autonomous driving via discrete diffusion},
  author={Zhang, Lunjun and Xiong, Yuwen and Yang, Ze and Casas, Sergio and Hu, Rui and Urtasun, Raquel},
  journal={ICLR},
  year={2024}
}

@inproceedings{VIDAR,
  title={Visual point cloud forecasting enables scalable autonomous driving},
  author={Yang, Zetong and Chen, Li and Sun, Yanan and Li, Hongyang},
  booktitle={CVPR},
  year={2024}
}

@inproceedings{ultralidar,
  title={Learning compact representations for lidar completion and generation},
  author={Xiong, Yuwen and Ma, Wei-Chiu and Wang, Jingkang and Urtasun, Raquel},
  booktitle={CVPR},
  year={2023}
}

@inproceedings{4docc,
  title={Point cloud forecasting as a proxy for 4d occupancy forecasting},
  author={Khurana, Tarasha and Hu, Peiyun and Held, David and Ramanan, Deva},
  booktitle={CVPR},
  year={2023}
}

@inproceedings{occworld,
  title={Occworld: Learning a 3d occupancy world model for autonomous driving},
  author={Zheng, Wenzhao and Chen, Weiliang and Huang, Yuanhui and Zhang, Borui and Duan, Yueqi and Lu, Jiwen},
  booktitle={ECCV},
  year={2024},
}

@article{renderworld,
  title={RenderWorld: World Model with Self-Supervised 3D Label},
  author={Yan, Ziyang and Dong, Wenzhen and Shao, Yihua and Lu, Yuhang and Haiyang, Liu and Liu, Jingwen and Wang, Haozhe and Wang, Zhe and Wang, Yan and Remondino, Fabio and others},
  journal={arXiv preprint},
  year={2024}
}

@article{occllama,
  title={Occllama: An occupancy-language-action generative world model for autonomous driving},
  author={Wei, Julong and Yuan, Shanshuai and Li, Pengfei and Hu, Qingda and Gan, Zhongxue and Ding, Wenchao},
  journal={arXiv preprint},
  year={2024}
}

@article{dfit-occworld,
  title={An Efficient Occupancy World Model via Decoupled Dynamic Flow and Image-assisted Training},
  author={Zhang, Haiming and Xue, Ying and Yan, Xu and Zhang, Jiacheng and Qiu, Weichao and Bai, Dongfeng and Liu, Bingbing and Cui, Shuguang and Li, Zhen},
  journal={arXiv preprint},
  year={2024}
}

@inproceedings{driveworld,
  title={Driveworld: 4d pre-trained scene understanding via world models for autonomous driving},
  author={Min, Chen and Zhao, Dawei and Xiao, Liang and Zhao, Jian and Xu, Xinli and Zhu, Zheng and Jin, Lei and Li, Jianshu and Guo, Yulan and Xing, Junliang and others},
  booktitle={CVPR},
  year={2024}
}

@article{drive-occworld,
  title={Driving in the occupancy world: Vision-centric 4d occupancy forecasting and planning via world models for autonomous driving},
  author={Yang, Yu and Mei, Jianbiao and Ma, Yukai and Du, Siliang and Chen, Wenqing and Qian, Yijie and Feng, Yuxiang and Liu, Yong},
  journal={AAAI},
  year={2025}
}

@article{dome,
  title={Dome: Taming diffusion model into high-fidelity controllable occupancy world model},
  author={Gu, Songen and Yin, Wei and Jin, Bu and Guo, Xiaoyang and Wang, Junming and Li, Haodong and Zhang, Qian and Long, Xiaoxiao},
  journal={arXiv preprint},
  year={2024}
}

@inproceedings{worldliao,
  title={$I^{2}$-World: Intra-Inter Tokenization for Efficient Dynamic 4D Scene Forecasting},
  author={Liao, Zhimin and Wei, Ping and Zhang, Ruijie and Chen, Shuaijia and Wang, Haoxuan and Ren, Ziyang},
  booktitle={ICCV},
  year={2025}
}

@article{occsora,
  title={Occsora: 4d occupancy generation models as world simulators for autonomous driving},
  author={Wang, Lening and Zheng, Wenzhao and Ren, Yilong and Jiang, Han and Cui, Zhiyong and Yu, Haiyang and Lu, Jiwen},
  journal={arXiv preprint},
  year={2024}
}

@article{urbandiff,
  title={Urban scene diffusion through semantic occupancy map},
  author={Zhang, Junge and Zhang, Qihang and Zhang, Li and Kompella, Ramana Rao and Liu, Gaowen and Zhou, Bolei},
  journal={arXiv preprint arXiv:2403.11697},
  year={2024}
}

@article{x-scene,
  title={X-Scene: Large-Scale Driving Scene Generation with High Fidelity and Flexible Controllability},
  author={Yang, Yu and Liang, Alan and Mei, Jianbiao and Ma, Yukai and Liu, Yong and Lee, Gim Hee},
  journal={arXiv preprint arXiv:2506.13558},
  year={2025}
}

@inproceedings{nuscenes,
author = {Caesar, Holger and Bankiti, Varun and Lang, Alex H. and Vora, Sourabh and Liong, Venice Erin and Xu, Qiang and Krishnan, Anush and Pan, Yu and Baldan, Giancarlo and Beijbom, Oscar},
title = {nuScenes: A Multimodal Dataset for Autonomous Driving},
booktitle = {CVPR},
year = {2020}
}

@inproceedings{occ3d,
  title={Occ3d: A large-scale 3d occupancy prediction benchmark for autonomous driving},
  author={Tian, Xiaoyu and Jiang, Tao and Yun, Longfei and Mao, Yucheng and Yang, Huitong and Wang, Yue and Wang, Yilun and Zhao, Hang},
  booktitle={NeurIPS},
  year={2023}
}

@inproceedings{waymo,
  title={Scalability in perception for autonomous driving: Waymo open dataset},
  author={Sun, Pei and Kretzschmar, Henrik and Dotiwalla, Xerxes and Chouard, Aurelien and Patnaik, Vijaysai and Tsui, Paul and Guo, James and Zhou, Yin and Chai, Yuning and Caine, Benjamin and others},
  booktitle={CVPR},
  year={2020}
}

@inproceedings{magicdrive,
  title={Magicdrive: Street view generation with diverse 3d geometry control},
  author={Gao, Ruiyuan and Chen, Kai and Xie, Enze and Hong, Lanqing and Li, Zhenguo and Yeung, Dit-Yan and Xu, Qiang},
  booktitle={ICLR},
  year={2024}
}

@inproceedings{magicdrive-v2,
  title={MagicDrive-V2: High-resolution long video generation for autonomous driving with adaptive control},
  author={Gao, Ruiyuan and Chen, Kai and Xiao, Bo and Hong, Lanqing and Li, Zhenguo and Xu, Qiang},
  booktitle={ICCV},
  year={2025}
}

@article{worldmodel_survey,
  title={3D and 4D world modeling: A survey},
  author={Kong, Lingdong and Yang, Wesley and Mei, Jianbiao and Liu, Youquan and Liang, Ao and Zhu, Dekai and Lu, Dongyue and Yin, Wei and Hu, Xiaotao and Jia, Mingkai and others},
  journal={arXiv preprint arXiv:2509.07996},
  year={2025}
}

@inproceedings{semantickitti,  
title={SemanticKITTI: A Dataset for Semantic Scene Understanding of LiDAR Sequences}, 
author={Behley, Jens and Garbade, Martin and Milioto, Andres and Quenzel, Jan and Behnke, Sven and Stachniss, Cyrill and Gall, Jurgen}, 
year={2019}, 
booktitle={ICCV}, 
}

@inproceedings{scene_as_occupancy,
  title={Scene as occupancy},
  author={Tong, Wenwen and Sima, Chonghao and Wang, Tai and Chen, Li and Wu, Silei and Deng, Hanming and Gu, Yi and Lu, Lewei and Luo, Ping and Lin, Dahua and others},
  booktitle={Proceedings of the IEEE/CVF International Conference on Computer Vision},
  pages={8406--8415},
  year={2023}
}

@inproceedings{bevformer,
  title={Bevformer: Learning bird’s-eye-view representation from multi-camera images via spatiotemporal transformers},
  author={Li, Zhiqi and Wang, Wenhai and Li, Hongyang and Xie, Enze and Sima, Chonghao and Lu, Tong and Qiao, Yu and Dai, Jifeng},
  booktitle={ECCV},
  year={2022},
}

@article{lagen,
  title={LaGen: Towards Autoregressive LiDAR Scene Generation},
  author={Zhou, Sizhuo and Jia, Xiaosong and Zhang, Fanrui and Li, Junjie and Zhang, Juyong and Feng, Yukang and Sun, Jianwen and Wong, Songbur and You, Junqi and Yan, Junchi},
  journal={arXiv preprint},
  year={2025}
}

@article{VAE,
  title={Auto-encoding variational bayes},
  author={Kingma, Diederik P and Welling, Max},
  journal={arXiv preprint arXiv:1312.6114},
  year={2013}
}

@inproceedings{stcocc,
  title={Stcocc: Sparse spatial-temporal cascade renovation for 3d occupancy and scene flow prediction},
  author={Liao, Zhimin and Wei, Ping and Chen, Shuaijia and Wang, Haoxuan and Ren, Ziyang},
  booktitle={CVPR},
  year={2025}
}

@article{ddpm,
  title={Denoising diffusion probabilistic models},
  author={Ho, Jonathan and Jain, Ajay and Abbeel, Pieter},
  journal={NIPS},
  year={2020}
}

@inproceedings{hermes,
  title={Hermes: A unified self-driving world model for simultaneous 3d scene understanding and generation},
  author={Zhou, Xin and Liang, Dingkang and Tu, Sifan and Chen, Xiwu and Ding, Yikang and Zhang, Dingyuan and Tan, Feiyang and Zhao, Hengshuang and Bai, Xiang},
  booktitle={ICCV},
  year={2025}
}

@article{uniscene2,
  title={Scaling up occupancy-centric driving scene generation: Dataset and method},
  author={Li, Bohan and Jin, Xin and Zhu, Hu and Liu, Hongsi and Li, Ruikai and Guo, Jiazhe and Cai, Kaiwen and Ma, Chao and Jin, Yueming and Zhao, Hao and others},
  journal={IEEE Transactions on Pattern Analysis and Machine Intelligence},
  year={2026},
  publisher={IEEE}
}

@inproceedings{geniedrive,
  title={Geniedrive: Towards physics-aware driving world model with 4d occupancy guided video generation},
  author={Yang, Zhenya and Liu, Zhe and Lu, Yuxiang and Hou, Liping and Miao, Chenxuan and Peng, Siyi and Feng, Bailan and Bai, Xiang and Zhao, Hengshuang},
  booktitle={CVPR},
  year={2026}
}

@inproceedings{deformabel-attn,
  title={Vision transformer with deformable attention},
  author={Xia, Zhuofan and Pan, Xuran and Song, Shiji and Li, Li Erran and Huang, Gao},
  booktitle={CVPR},
  year={2022}
}

@article{bevworld,
  title={Bevworld: A multimodal world model for autonomous driving via unified bev latent space},
  author={Zhang, Yumeng and Gong, Shi and Xiong, Kaixin and Ye, Xiaoqing and Tan, Xiao and Wang, Fan and Huang, Jizhou and Wu, Hua and Wang, Haifeng},
  year={2024}
}

@article{flowmatching,
  title={Flow matching for generative modeling},
  author={Lipman, Yaron and Chen, Ricky TQ and Ben-Hamu, Heli and Nickel, Maximilian and Le, Matt},
  journal={arXiv preprint},
  year={2022}
}

@article{fbocc,
  title={Fb-occ: 3d occupancy prediction based on forward-backward view transformation},
  author={Li, Zhiqi and Yu, Zhiding and Austin, David and Fang, Mingsheng and Lan, Shiyi and Kautz, Jan and Alvarez, Jose M},
  journal={arXiv preprint},
  year={2023}
}

@inproceedings{lovaszloss,
  title={The lov{\'a}sz-softmax loss: A tractable surrogate for the optimization of the intersection-over-union measure in neural networks},
  author={Berman, Maxim and Triki, Amal Rannen and Blaschko, Matthew B},
  booktitle={CVPR},
  year={2018}
}

@inproceedings{fpd,
  title={3d point cloud generative adversarial network based on tree structured graph convolutions},
  author={Shu, Dong Wook and Park, Sung Woo and Kwon, Junseok},
  booktitle={ICCV},
  year={2019}
}

@inproceedings{MMD,
  title={Learning representations and generative models for 3d point clouds},
  author={Achlioptas, Panos and Diamanti, Olga and Mitliagkas, Ioannis and Guibas, Leonidas},
  booktitle={ICML},
  year={2018}
}

@inproceedings{dit,
  title={Scalable diffusion models with transformers},
  author={Peebles, William and Xie, Saining},
  booktitle={2023 IEEE/CVF International Conference on Computer Vision (ICCV)},
  pages={4172--4182},
  year={2023},
  organization={IEEE}
}
}

% Working supplementary manuscript, integrated from the author-approved draft.
% Merged figure layout approved and integrated on 2026-11-01.
% Figure metadata (model/time, RGB source, JC scope/color key) remain pending.
% Approval covers layout, not unconfirmed experimental metadata.
% See temp/supplement-figure-integration-20261101/revision-notes.md.
% JC definition and author-confirmed reference thresholds are integrated.
% Training coefficients and forecasting supervision are integrated.
% Shared edge weighting is aligned with the main range-loss formula.
% See temp/loss-edge-consistency-20261031/revision-notes.md and
% temp/supplement-integration-20261031/pending-work.md.
\clearpage
\setcounter{page}{1}
\maketitlesupplementary
\raggedbottom
\setlength{\dbltextfloatsep}{12pt plus 2pt minus 2pt}
\makeatletter
\setlength{\@dblfptop}{0pt}
\makeatother
\setcounter{section}{0}
\renewcommand{\thesection}{\Alph{section}}
\renewcommand{\thesubsection}{\thesection.\arabic{subsection}}
\renewcommand{\theHsection}{supp.\Alph{section}}
\setcounter{table}{0}
\renewcommand{\thetable}{A\arabic{table}}
\renewcommand{\theHtable}{supp.\arabic{table}}
\setcounter{figure}{0}
\renewcommand{\thefigure}{A\arabic{figure}}
\renewcommand{\theHfigure}{supp.\arabic{figure}}
\setcounter{equation}{0}
\renewcommand{\theequation}{A\arabic{equation}}
\renewcommand{\theHequation}{supp.\arabic{equation}}
\makeatletter
\setlength{\@fptop}{0pt}
\makeatother

\noindent
This supplementary material is organized as follows:
\begin{itemize}[leftmargin=*, topsep=3pt, itemsep=2pt, parsep=0pt]
    \item \textbf{Implementation details (Sec.~\ref{sec:supp-implementation}).}
    We provide details of the data representation, model training, conditioning, downstream video generation, and baseline designs.
    \item \textbf{Evaluation details (Sec.~\ref{sec:supp-evaluation}).}
    We specify the evaluation protocol, define Joint Consistency, and describe the modality-specific metrics and efficiency measurements.
    \item \textbf{Additional results (Sec.~\ref{sec:supp-additional-results}).}
    We present 12-frame forecasting results and qualitative examples of Joint Consistency and scene structure.
    \item \textbf{Scope and limitations (Sec.~\ref{sec:supp-limitations}).}
    We discuss reverse-motion failures and the need for downstream task validation.
\end{itemize}

\section{Implementation Details}
\label{sec:supp-implementation}

\subsection{Data Representation and Preprocessing}
\label{sec:supp-data}

\paragraph{Spatial representation.}
We pair each nuScenes LiDAR scan with the occupancy annotation of the corresponding frame.
Occupancy is represented in the ego frame, while range projection is performed in the LiDAR sensor frame.
We align the back-projected range samples with the occupancy grid using the calibrated LiDAR-to-ego transformation.
Table~\ref{tab:supp-data-settings} summarizes the spatial and temporal settings.

\paragraph{LiDAR preprocessing.}
We use individual scans without multi-sweep accumulation.
Points within the ego-vehicle exclusion region are removed, and the remaining points are restricted to the ego-frame spatial bounds.
We then project them onto a spherical range image, retaining the nearest return at each pixel.
Projection indices outside the image are clipped to its boundaries.
Pixels without a return are set to zero. The tokenizer treats only depths satisfying $d_{\min}<d<d_{\max}$ as valid.

\paragraph{Temporal setting.}
The model observes the current frame and four past frames over 2\,s, then predicts six future frames at 0.5\,s intervals.
Future ego motion is provided as a condition, whereas future occupancy and LiDAR are used only as supervision and evaluation references.

\begin{table}[t]
    \centering
    \small
    \setlength{\tabcolsep}{4pt}
    \renewcommand{\arraystretch}{1.12}
    \begin{tabularx}{\columnwidth}{@{}>{\raggedright\arraybackslash}p{0.43\columnwidth}>{\raggedright\arraybackslash}X@{}}
        \toprule
        Setting & Value \\
        \midrule
        Ego-frame bounds (m) &
        \makecell[l]{$[-40,40)\times[-40,40)$\\$\times[-1,5.4)$} \\
        Occupancy voxel size & $0.4\times0.4\times0.4$ m \\
        Occupancy grid $(N_x,N_y,N_z)$ & $200\times200\times16$ \\
        Occupancy labels & 18 (free-space ID 17) \\
        Range-image size $(H_r,W_r)$ & $32\times1024$ \\
        Vertical projection bounds & $-30^\circ,\,+10^\circ$ \\
        Depth thresholds $(d_{\min},d_{\max})$ & $(1.45,80)$ m \\
        Ego exclusion region (m) &
        \makecell[l]{$[-1.542,3.042]$\\$\times[-0.925,0.925]$} \\
        Sampling interval & 0.5 s \\
        Observed / predicted frames & 5 / 6 \\
        \bottomrule
    \end{tabularx}
    \caption{Spatial and temporal settings for the six-frame nuScenes task. Spatial bounds follow $(x,y,z)$, and the ego exclusion region uses sensor-frame $(x,y)$ coordinates. Observed frames include the current frame.}
    \label{tab:supp-data-settings}
\end{table}

\subsection{LO-VAE}
\label{sec:supp-vae}

\paragraph{Shared representation.}
LO-VAE maps occupancy and LiDAR into a shared BEV latent.
Occupancy embeddings are collapsed along height, while normalized LiDAR range and validity features are aggregated into voxels and arranged in BEV.
Modality-specific convolutional stems process these representations before a shared encoder fuses their concatenated features.
A shared decoder then recovers scene features for both output heads.
Table~\ref{tab:supp-model-settings} summarizes the architecture and sampling settings.

\paragraph{Ray-based decoding.}
The LiDAR head lifts the shared BEV features into a 3D feature volume.
For each sensor ray, it samples this volume and the predicted occupancy context at 192 candidate depths, uniformly spaced in normalized log-depth.
We add an occupied-probability prior to the depth scores and use a separate class for empty rays.
Training targets assign each valid return to the nearest candidate in metric depth.
At inference, we select the most probable class and convert non-empty predictions to metric 3D points along their sensor rays.

\paragraph{Reconstruction losses.}
Occupancy reconstruction combines sigmoid focal loss and Lovasz loss with coefficients 10 and 1.
The focal loss uses $\gamma=2$, $\alpha=0.25$, and a spatial weight $1+\rho/\rho_{\max}$, where $\rho$ is the BEV distance to the grid center and $\rho_{\max}$ its maximum.
We sum focal terms over classes before averaging over labeled voxels, and compute Lovasz loss over the batch for classes present in the targets.
Both losses exclude unknown labels (ID 255).

LiDAR classification uses base weights of 1 for valid returns and 0.1 for empty rays.
Depth regression uses Smooth L1 loss with coefficient 2 and unit transition threshold, only on valid rays.
Its prediction is the expected normalized log-depth $u(d)=2\log(d+1)/\log(d_{\max}+1)-1$ over the non-empty candidates. The empty-ray class does not enter this expectation.
Classification is averaged over all rays, whereas regression is averaged only over valid rays.
The overall occupancy and LiDAR coefficients are both 1, and the KL coefficient is $10^{-6}$.
The KL term is the summed latent divergence divided by the latent width (50), averaged across samples.

Both classification and regression emphasize depth discontinuities.
Depth edges occur between horizontally or vertically adjacent valid rays whose normalized log-depth difference exceeds 0.08, with wrap-around adjacency horizontally.
Both rays at an edge receive a factor of 3, compared with 1 elsewhere.
The edge weights $e_p$ are normalized to mean one over valid rays in the batch.
Classification uses $w_p$, the product of $e_p$ and the base weight, while Smooth L1 regression uses $e_p$ directly. Empty-ray weights are unchanged.

\paragraph{Cross-modal gradients.}
LiDAR supervision updates the shared scene features but does not backpropagate through the occupancy probabilities supplied to the LiDAR head.
LiDAR gradients to the shared features are scaled by 0.25.

\begin{table}[t]
    \centering
    \small
    \setlength{\tabcolsep}{4pt}
    \renewcommand{\arraystretch}{1.12}
    \begin{tabularx}{\columnwidth}{@{}>{\raggedright\arraybackslash}X>{\raggedright\arraybackslash}p{0.32\columnwidth}@{}}
        \toprule
        Setting & Value \\
        \midrule
        \multicolumn{2}{@{}l}{LO-VAE} \\
        Occupancy embedding width & 16 \\
        Stem width per modality & 64 \\
        Encoder stage widths & 64, 128, 256 \\
        Spatial downsampling factor & 4 \\
        Latent size & $50\times50\times16$ \\
        Shared decoded feature width & 64 \\
        Depth candidates / empty class & 192 / 1 \\
        \midrule
        \multicolumn{2}{@{}l}{LO-Former} \\
        Hidden width / attention heads & 512 / 8 \\
        Context blocks / flow blocks & 8 / 4 \\
        Latent patch size & $1\times1$ \\
        MLP expansion ratio & 4 \\
        Attention / projection dropout & 0.1 / 0.1 \\
        Euler sampling steps & 4 \\
        \bottomrule
    \end{tabularx}
    \caption{Architecture and sampling settings for LO-VAE and LO-Former. Latent sizes list spatial dimensions followed by channel width. Encoder widths follow stage order.}
    \label{tab:supp-model-settings}
\end{table}

\subsection{LO-Former}
\label{sec:supp-former}
\paragraph{Training stages.}
We first train LO-VAE to reconstruct paired observations.
With LO-VAE fixed, we train LO-Former in its latent space, progressing from two-frame to six-frame prediction.
For 12-frame prediction over 6\,s, we further fine-tune the six-frame model.
Table~\ref{tab:supp-training-settings} lists the settings for these stages.

\paragraph{Optimization.}
All stages use AdamW with $\beta_1=0.9$, weight decay $0.01$, and a 500-iteration linear learning-rate warmup.
LO-VAE follows a cosine schedule with a minimum learning rate of $10^{-6}$.
LO-Former uses step decay, reducing the learning rate by a factor of $0.1$ at each listed epoch.

\begin{table}[t]
    \centering
    \small
    \setlength{\tabcolsep}{2.5pt}
    \renewcommand{\arraystretch}{1.12}
    \begin{tabularx}{\columnwidth}{@{}>{\raggedright\arraybackslash}p{0.32\columnwidth}*{4}{>{\centering\arraybackslash}X}@{}}
        \toprule
        Setting & LO-VAE & \multicolumn{3}{c@{}}{LO-Former} \\
        \cmidrule(l){3-5}
        & & 2-frame & 6-frame & 12-frame \\
        \midrule
        Learning rate & $10^{-4}$ & $5\times10^{-4}$ & $10^{-4}$ & $10^{-5}$ \\
        Total batch size & 16 & 8 & 8 & 8 \\
        Training epochs & 40 & 72 & 20 & 8 \\
        AdamW $\beta_2$ & 0.95 & 0.999 & 0.999 & 0.999 \\
        Gradient clipping norm & 1 & 15 & 15 & 15 \\
        Initial warmup LR ratio & $10^{-2}$ & $10^{-3}$ & $10^{-3}$ & $10^{-3}$ \\
        Learning-rate schedule & Cosine & Step & Step & Step \\
        Decay epochs & -- & 40, 56 & 10, 14 & 4, 6 \\
        \bottomrule
    \end{tabularx}
    \caption{Training settings for LO-VAE and the successive LO-World-T forecasting stages on nuScenes. The 12-frame stage fine-tunes the six-frame model. Batch sizes are totals across eight GPUs. Warmup ratios are relative to the listed learning rates (LR).}
    \label{tab:supp-training-settings}
\end{table}

\paragraph{Forecasting objective.}
In addition to latent flow matching, we supervise decoded LiDAR predictions:
\begin{equation}
    \mathcal{L}_{\mathrm{forecast}}
    =\mathcal{L}_{\mathrm{FM}}+0.05\,\mathcal{L}_{\mathrm{LiDAR}}^{\mathrm{aux}}.
    \label{eq:supp-forecast-objective}
\end{equation}
The auxiliary term applies the LiDAR classification and regression losses to selected clean-latent predictions $\hat{\mathbf{x}}_0=\mathbf{x}_{\sigma}-\sigma\hat{\mathbf{v}}_{\psi}$ decoded by LO-VAE.
It updates LO-Former through the fixed decoder, without additional occupancy reconstruction or KL losses.

Flow matching supervises every future latent, averaging the squared velocity error over the batch, frames, and latent elements.
The auxiliary LiDAR loss uses both outputs in two-frame training (0.5 and 1\,s), but only the last two outputs in six-frame training (2.5 and 3\,s).
During 12-frame fine-tuning, the first and second six-frame chunks are supervised alternately with equal loss coefficients.
The second chunk is conditioned on a sampled first chunk held fixed during this training step.
Auxiliary LiDAR supervision remains on the last two frames of the supervised chunk, corresponding to 5.5 and 6\,s for the second chunk.
These loss settings are shared by LO-World-T and LO-World-TBM.

\paragraph{Training sequences.}
When the four-frame history is incomplete, we repeat the first observation.
Unavailable future targets are replaced with the last observation and assigned weight 0.1 in both forecasting losses, without renormalizing the remaining terms.
Evaluation instead requires complete future windows (Sec.~\ref{sec:supp-time-eval}).

\paragraph{Sampling.}
LO-Former jointly generates six future latents from Gaussian noise using four Euler steps and a single encoding of the history.
LO-VAE decodes the resulting latents into occupancy grids and LiDAR range images.

\paragraph{Long-horizon rollout.}
For a 6\,s rollout, we generate two consecutive six-frame chunks.
Before the second chunk, we re-encode the updated history using predictions from the first chunk rather than future ground-truth observations.
Generation remains joint within each chunk.

\subsection{Conditioning and Cross-Dataset Evaluation}
\label{sec:supp-conditions}

\paragraph{Conditioning variants.}
Both LO-World-T and LO-World-TBM forecast from paired historical occupancy and LiDAR, with relative ego rotation and translation as frame-wise conditions.
LO-World-TBM also uses box- and map-derived layouts as spatial conditions.
These layouts are supplied externally rather than predicted by LO-World.
For 12-frame prediction, LO-World-TBM is fine-tuned from its six-frame counterpart.
It follows the 12-frame settings in Table~\ref{tab:supp-training-settings}, with a tenfold learning rate for the layout embedding and adapter.

\paragraph{Camera-derived occupancy.}
In the STC setting, camera-predicted occupancy replaces the annotated occupancy input, while LiDAR remains an observed modality.
Occupancy forecasts are still evaluated against the ground-truth annotations.

\paragraph{Waymo evaluation.}
We evaluate both zero-shot transfer to Waymo and adaptation using 50 target-domain training sequences on the same 202 validation sequences described in the main paper.
The LiDAR output uses a $64\times2048$ range image with vertical projection bounds of $-17.734^\circ$ and $2.274^\circ$.

\subsection{Downstream Video Generation}
\label{sec:supp-downstream}

\paragraph{Occupancy conditioning.}
To provide occupancy conditions for video generation, we apply LO-World-TBM zero-shot at 12\,Hz using the supplied 12\,Hz ego trajectories and box- and map-derived layouts.
This changes the prediction frequency without further training the occupancy model, unlike the 12-frame fine-tuning above.
The predicted occupancy sequences are fed to GenieDrive~\cite{geniedrive} following its occupancy-input setup.
The 12\,Hz ground-truth occupancy reference comes from UniScene~\cite{uniscene}.

\paragraph{Video training and evaluation.}
We fine-tune the video model using the first 81 frames of each training scene and evaluate on the first 81 frames of each validation scene.
FVD, mIoU, and mAP are computed following the evaluation protocol of MagicDrive~\cite{magicdrive}.

\subsection{Modality-Specific Baselines}
\label{sec:supp-baselines}

\paragraph{Single-modality autoencoders.}
The separate-latent baselines in Table~\ref{table:ablation_study_unified_vae} of the main paper use single-modality variants of LO-VAE.
The occupancy-only variant follows the same encoder and decoder architecture as LO-VAE.
The LiDAR-only variant retains the LiDAR pathway but removes the occupancy-context skip connection to the LiDAR decoder.

\paragraph{Token concatenation (a).}
The occupancy and LiDAR latent tokens are concatenated into a single sequence before being processed by LO-Former.
This differs from LO-World, which forecasts a shared latent representation rather than separately encoded modality tokens.

\paragraph{Cross-modal attention (b).}
The two modality-specific sequences remain separate and are processed in parallel, exchanging information through cross-modal attention in both the context and flow blocks of LO-Former.

\section{Evaluation Details}
\label{sec:supp-evaluation}

\subsection{Temporal Protocol and Aggregation}
\label{sec:supp-time-eval}

We evaluate nuScenes sequences with four available historical frames and a complete ground-truth future window.
A missing future frame excludes the sequence from evaluation at every horizon.
The six- and 12-frame settings require complete 3\,s and 6\,s windows, respectively, and therefore use different subsets of the same validation split.

The current-frame reconstruction is kept separate from future prediction.
For occupancy mIoU, non-empty IoU, and Chamfer Distance (CD), the main-table average uses the three reported horizons:
\begin{equation}
    \operatorname{Avg.}(m)=\frac{m(1\,\mathrm{s})+m(2\,\mathrm{s})+m(3\,\mathrm{s})}{3}.
    \label{eq:supp-average}
\end{equation}
This average is computed before rounding and does not include all six predicted frames.
The 12-frame results in Table~\ref{tab:supp-temporal-results} are reported separately at each horizon.

\subsection{Occupancy and LiDAR Metrics}
\label{sec:supp-modality-eval}

\paragraph{Occupancy.}
We compute occupancy scores from voxel-level confusion counts pooled across samples at each prediction horizon, excluding unknown voxels (label 255).
The label space contains 17 non-free categories and one free-space category.
Non-empty IoU merges all non-free labels into one occupied category and computes $\mathrm{TP}/(\mathrm{TP}+\mathrm{FP}+\mathrm{FN})$ from binary occupied/free decisions.
It therefore measures geometry without penalizing semantic confusions among occupied categories.
Semantic mIoU averages the non-free class IoUs after excluding all zero-IoU classes, including those present in ground truth, and is set to zero if no class remains.
Both scores are reported as percentages and are based on pooled counts rather than mean sample-level IoUs.

\paragraph{LiDAR geometry.}
We compare predicted and reference point clouds in the LiDAR sensor frame using 3D coordinates only, without intensity.
The reference cloud is reconstructed from the projected range image, not the raw scan.
Using the angular grid in Table~\ref{tab:supp-data-settings}, we retain back-projected depths strictly between 1.45 and 80\,m.
All valid back-projected points are used in nuScenes evaluation, without further subsampling.
For the predicted and reference point sets $\widehat{\mathcal{P}}$ and $\mathcal{P}$ used in evaluation, CD is
\begin{equation}
    \begin{aligned}
    \operatorname{CD}(\widehat{\mathcal{P}},\mathcal{P})
    ={}&\frac{1}{|\widehat{\mathcal{P}}|}
    \sum_{\hat p\in\widehat{\mathcal{P}}}\min_{p\in\mathcal{P}}\|\hat p-p\|_2^2\\
    &+\frac{1}{|\mathcal{P}|}
    \sum_{p\in\mathcal{P}}\min_{\hat p\in\widehat{\mathcal{P}}}\|p-\hat p\|_2^2.
    \end{aligned}
    \label{eq:supp-cd}
\end{equation}
CD is the sum of the two directional mean squared distances, with no factor of $1/2$ or square root.
With coordinates in meters, its units are $\mathrm{m}^2$.
We average cloud-pair scores equally across samples at each horizon.
% The bundled Chamfer kernel confirms squared distances and the reduction above.
% At 32 x 1024, the point-count guard cannot activate. This does not imply
% evaluation against every raw return. Empty-set handling and baseline
% protocol equivalence still require final-run confirmation.

% Queue full-width figures ahead of their discussion so text can flow below them.
\begin{figure*}[t]
    \centering
    \includegraphics[width=\textwidth]{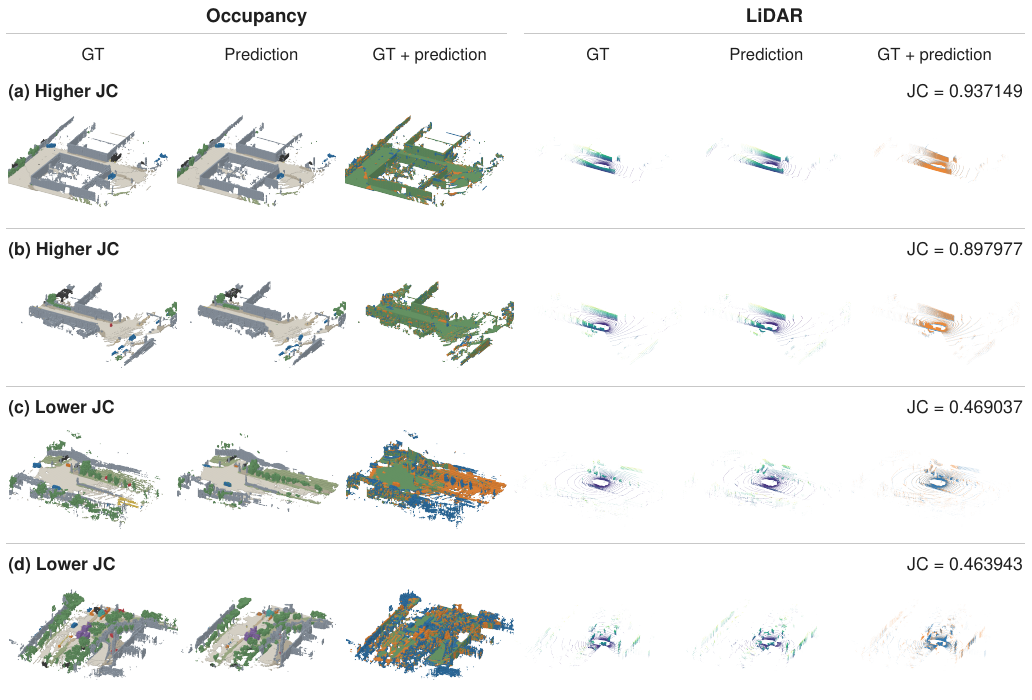}
    \caption{Examples with higher and lower JC. Each row pairs occupancy and LiDAR views from one example. For each modality, we show ground truth (GT), prediction, and their superimposition. The associated JC score is shown at the right.}
    \label{fig:supp-jc-examples}
\end{figure*}

\subsection{Joint Consistency}
\label{sec:supp-jc}

Joint Consistency (JC) measures whether occupancy and LiDAR predictions jointly recover the same ground-truth geometry.
It combines joint surface recall with occupancy precision under a reference determined solely by ground truth.
JC treats all occupied categories alike and measures geometric rather than semantic agreement.

\paragraph{Ground-truth reference.}
We use the angular grid and depth limits in Table~\ref{tab:supp-data-settings}, retaining the nearest original LiDAR return in each angular bin.
Reference returns are restricted to the occupancy region and transformed to the ego frame.
Let $p_i$ denote a raw ground-truth return and $d(p_i,v)$ its Euclidean distance to the axis-aligned box of voxel $v$, rather than its center.
We assign $p_i$ the nearest ground-truth occupied voxel $v_i$, breaking ties by lexicographic voxel order.
With support tolerance $\tau=0.4$\,m, the return is retained only if $d(p_i,v_i)\leq\tau$ and no unknown voxel lies within distance $\tau$.
These retained returns and their assigned voxels form a fixed reference shared by all predictions.

\paragraph{Evaluation region.}
The local region $\mathcal{D}_{\mathrm{local}}$ contains voxel boxes within radius $r=0.8$\,m of any retained return.
The ray region $\mathcal{D}_{\mathrm{ray}}$ contains voxels traversed along the original ground-truth ray to each retained return, starting at the minimum valid depth $d_{\min}=1.45$\,m.
The rays stop at their returns without extending into unobserved space behind them.
We take the union of these two regions and exclude unknown ground-truth voxels:
\begin{equation}
    \mathcal{D}
    =(\mathcal{D}_{\mathrm{local}}\cup\mathcal{D}_{\mathrm{ray}})
    \cap\mathcal{V}_{\mathrm{known}},
    \label{eq:supp-jc-domain}
\end{equation}
where $\mathcal{V}_{\mathrm{known}}$ is the set of voxels with known occupancy labels.
Voxels along a ray retain their ground-truth occupied or free labels rather than being uniformly treated as free.

\paragraph{Joint surface recall.}
Predicted scans use the same angular binning and depth limits, without an additional spatial crop.
For each retained reference return, we select the predicted return $\hat p_i$ from the same angular bin and transform it to the ego frame.
Let $\tilde p_i$ be the ground-truth return reconstructed from its range-image depth along the bin-center ray.
To account for angular discretization, we use the matching tolerance
\begin{equation}
    \epsilon_i=\tau+\|\tilde p_i-p_i\|_2.
    \label{eq:supp-jc-tolerance}
\end{equation}
A joint match requires a valid predicted return satisfying all three conditions:
\begin{equation}
    \|\hat p_i-p_i\|_2\leq\epsilon_i,\quad
    \hat o(v_i)=1,\quad
    d(\hat p_i,v_i)\leq\epsilon_i,
    \label{eq:supp-jc-match}
\end{equation}
where $\hat o(v)=1$ denotes predicted occupancy.
Thus, both modalities must support the same fixed reference voxel, and the LiDAR return must also match the observed surface.
A missing return is a failed match, not an excluded reference.

\paragraph{Score and aggregation.}
Let $N$ be the number of retained reference returns and $J$ the number of joint matches.
Within $\mathcal{D}$, let $\mathrm{TP}$ count predicted-occupied voxels labeled occupied in ground truth, and $\mathrm{FP}$ count predicted-occupied voxels labeled free.
Joint recall $R$, occupancy precision $P$, and JC are
\begin{equation}
    \begin{gathered}
        R=\frac{J}{N},\qquad
        P=\frac{\mathrm{TP}}{\mathrm{TP}+\mathrm{FP}},\\
        \mathrm{JC}=\frac{2PR}{P+R}.
    \end{gathered}
    \label{eq:supp-jc-score}
\end{equation}
We first sum these counts across evaluated samples and all six future frames, excluding the current-frame reconstruction, and then compute the score.
JC is therefore neither an average of frame-level scores nor the three-horizon average in Eq.~\eqref{eq:supp-average}.
We set $P=0$ when there are no predicted-occupied voxels in $\mathcal{D}$, and JC to zero when $P+R=0$.
If the pooled reference contains no retained returns, the score is undefined and is not reported.
Higher JC requires both joint recovery of observed surfaces and fewer false occupied predictions in the ground-truth-defined region.
It does not assess consistency outside that region.

\paragraph{Visual examples.}
Figure~\ref{fig:supp-jc-examples} provides visual examples, which we discuss in Sec.~\ref{sec:supp-qualitative}.

\subsection{Distributional Metrics}
\label{sec:supp-distribution-eval}

We assess distributional fidelity by comparing collections of predicted and reference point clouds, rather than individual prediction--reference pairs.

\paragraph{Point-cloud features.}
FPD compares the means and covariances of features from a pretrained PointNet.
Point coordinates are divided by 80 before feature extraction, producing one 1808-dimensional feature per cloud.

\paragraph{BEV distributions.}
B-MMD and B-JSD instead use $100\times100$ BEV histograms over $[-80,80]\times[-80,80]$\,m in the sensor-frame horizontal plane.
Only points with 3D distance strictly between 3 and 70\,m are retained for these histograms.
For B-MMD, each cloud's histogram is normalized to unit mass and compared using the RBF kernel
$k(\mathbf{a},\mathbf{b})=\exp(-2\|\mathbf{a}-\mathbf{b}\|_2^2)$.
We report squared MMD, using all within-set and cross-set histogram pairs, including self-pairs.
B-JSD instead pools point counts across clouds and then normalizes each aggregate histogram to unit mass.
We report the square root of the Jensen--Shannon divergence between these aggregate distributions, using natural logarithms.
Thus, B-MMD gives equal weight to each cloud, whereas the B-JSD aggregates weight clouds in proportion to their retained point counts.
The main table displays B-MMD scaled by $10^4$ and B-JSD scaled by $10^2$.

\paragraph{Temporal pooling.}
Each distributional score is computed once from all six future frames across evaluated samples, excluding current-frame reconstruction.
Each cloud contributes its own feature or histogram, without merging points across time.
This differs from averaging scores computed separately at 1, 2, and 3\,s.

\subsection{Efficiency Reporting}
\label{sec:supp-efficiency}

We measure throughput on one NVIDIA RTX 4090 with a batch size of one.
Model FPS includes current-frame encoding and latent forecasting, but excludes output decoding.
FPS-E2E additionally includes decoding occupancy and LiDAR for all six predicted future frames.
Both report generated future frames per second, rather than six-frame rollouts per second.
We sum the durations of the included stages, synchronizing the GPU at stage boundaries and excluding warmup, data loading, and preprocessing.

\section{Additional Results}
\label{sec:supp-additional-results}

\subsection{Long-Horizon Forecasting}
\label{sec:supp-temporal-results}

To examine how forecasting accuracy changes over longer horizons, we evaluate the 12-frame fine-tuned LO-World-T and LO-World-TBM models over 6\,s (Table~\ref{tab:supp-temporal-results}).
Requiring a complete 6\,s future window also reduces the evaluation subset, as described in Sec.~\ref{sec:supp-time-eval}.
Because both the models and the subsets differ from the short-horizon setting, the first six horizons are not directly comparable to the main-table results.
Occupancy overlap decreases and CD increases with the prediction horizon for both variants.
LO-World-TBM maintains higher occupancy scores and lower CD throughout, consistent with the benefit of box and map conditioning over longer rollouts.
\begin{table}[!t]
    \centering
    \small
    \setlength{\tabcolsep}{2pt}
    \renewcommand{\arraystretch}{1.12}
    \begin{tabular*}{\columnwidth}{@{\extracolsep{\fill}}l*{6}{c}@{}}
        \toprule
        Time (s) & \multicolumn{3}{c}{LO-World-T} & \multicolumn{3}{c@{}}{LO-World-TBM} \\
        \cmidrule(lr){2-4}\cmidrule(l){5-7}
        & \shortstack{mIoU $\uparrow$\\(\%)} & \shortstack{IoU $\uparrow$\\(\%)} & \shortstack{CD $\downarrow$\\($\mathrm{m}^2$)} & \shortstack{mIoU $\uparrow$\\(\%)} & \shortstack{IoU $\uparrow$\\(\%)} & \shortstack{CD $\downarrow$\\($\mathrm{m}^2$)} \\
        \midrule
        0.5 & 60.02 & 64.58 & 0.3164 & 63.32 & 64.95 & 0.3066 \\
        1.0 & 53.56 & 61.72 & 0.4206 & 59.65 & 62.49 & 0.3896 \\
        1.5 & 48.89 & 59.31 & 0.5440 & 57.01 & 60.46 & 0.4829 \\
        2.0 & 45.15 & 57.14 & 0.6833 & 54.86 & 58.60 & 0.5817 \\
        2.5 & 42.00 & 55.02 & 0.7906 & 52.93 & 56.77 & 0.6526 \\
        3.0 & 39.22 & 52.97 & 0.9740 & 51.10 & 54.91 & 0.7628 \\
        \midrule
        3.5 & 35.96 & 50.28 & 1.2734 & 49.41 & 52.47 & 0.9560 \\
        4.0 & 34.01 & 48.65 & 1.4884 & 48.16 & 51.01 & 1.0986 \\
        4.5 & 32.30 & 47.15 & 1.7019 & 46.99 & 49.59 & 1.2609 \\
        5.0 & 30.70 & 45.62 & 1.9976 & 45.74 & 48.21 & 1.4621 \\
        5.5 & 29.12 & 43.99 & 2.1157 & 44.66 & 46.76 & 1.5186 \\
        6.0 & 27.48 & 42.38 & 2.4280 & 43.52 & 45.28 & 1.7239 \\
        \bottomrule
    \end{tabular*}
    \caption{Twelve-frame forecasting on nuScenes at 0.5\,s intervals. The divider after 3\,s separates the two six-frame chunks.}
    \label{tab:supp-temporal-results}
\end{table}

\subsection{Sampling Steps for Unimodal and Joint Forecasting}
\label{sec:supp-unimodal-sampling}

\begin{table}[t]
    \centering
    \small
    \setlength{\tabcolsep}{2pt}
    \renewcommand{\arraystretch}{1.12}
    \begin{tabular*}{\columnwidth}{@{\extracolsep{\fill}}c*{6}{c}@{}}
        \toprule
        Steps & \multicolumn{2}{c}{Occupancy-only} & \multicolumn{3}{c}{Joint} & \multicolumn{1}{c@{}}{LiDAR-only} \\
        \cmidrule(lr){2-3}\cmidrule(lr){4-6}\cmidrule(l){7-7}
        & \shortstack{mIoU $\uparrow$\\(\%)} & \shortstack{IoU $\uparrow$\\(\%)} & \shortstack{mIoU $\uparrow$\\(\%)} & \shortstack{IoU $\uparrow$\\(\%)} & \shortstack{CD $\downarrow$\\($\mathrm{m}^2$)} & \shortstack{CD $\downarrow$\\($\mathrm{m}^2$)} \\
        \midrule
        1 & 45.85 & 53.60 & 47.20 & 57.52 & 0.83 & 1.01 \\
        2 & 45.72 & 53.52 & 47.19 & 57.50 & 0.83 & 0.98 \\
        3 & 45.47 & 53.49 & 46.77 & 57.48 & 0.75 & 0.89 \\
        \rowcolor{aliceblue}
        4 & 45.28 & 53.47 & 46.01 & 57.46 & 0.70 & 0.82 \\
        8 & 42.51 & 51.46 & 43.80 & 55.33 & 0.68 & 0.81 \\
        \bottomrule
    \end{tabular*}
    \caption{Sampling-step ablation for unimodal and joint LO-World-T.}
    \label{tab:supp-unimodal-sampling}
\end{table}

To examine the sensitivity of sampling steps across generative formulations, we evaluate the occupancy-only, LiDAR-only, and joint variants of LO-World-T in \cref{tab:supp-unimodal-sampling} across Euler sampling steps $N \in \{1, 2, 3, 4, 8\}$. 

As the step count increases, both the occupancy-only and joint models maintain stable geometric IoU up to four steps before experiencing a slight decline at eight steps, while mIoU exhibits a gradual downward trend. Concurrently, increasing the number of steps steadily refines point cloud distributions, resulting in consistent Chamfer Distance reductions for both LiDAR-only and joint models. Crucially, joint forecasting outperforms the respective unimodal baselines across all tested sampling steps, simultaneously delivering higher semantic mIoU, superior geometric IoU, and lower Chamfer Distance under identical compute budgets. At our default setting of four steps, joint modeling improves geometric IoU by $3.99\%$ and decreases Chamfer Distance from $0.82$ to $0.70$. These results demonstrate that the bidirectional synergy between occupancy and LiDAR remains robust across varying numerical integration budgets.

\subsection{Module-Wise Complexity and Latency Analysis}

In \cref{tab:parameter}, we provide a detailed breakdown of the parameter count and inference latency for each individual module in LO-VAE and LO-Former.

% Keep the final text page balanced: long-horizon results on the left,
% qualitative interpretation and limitations on the right.

\subsection{Additional Visualizations}
\label{sec:supp-qualitative}

\paragraph{Joint Consistency examples.}
The higher-JC examples (a) and (b) in Fig.~\ref{fig:supp-jc-examples} show closer spatial agreement in the occupancy overlays.
The lower-JC examples (c) and (d) exhibit more visible discrepancies around foreground structures.
These examples illustrate the metric defined in Sec.~\ref{sec:supp-jc}, rather than provide a controlled comparison between model variants.

\begin{figure*}[p]
    \centering
    \includegraphics[width=\textwidth]{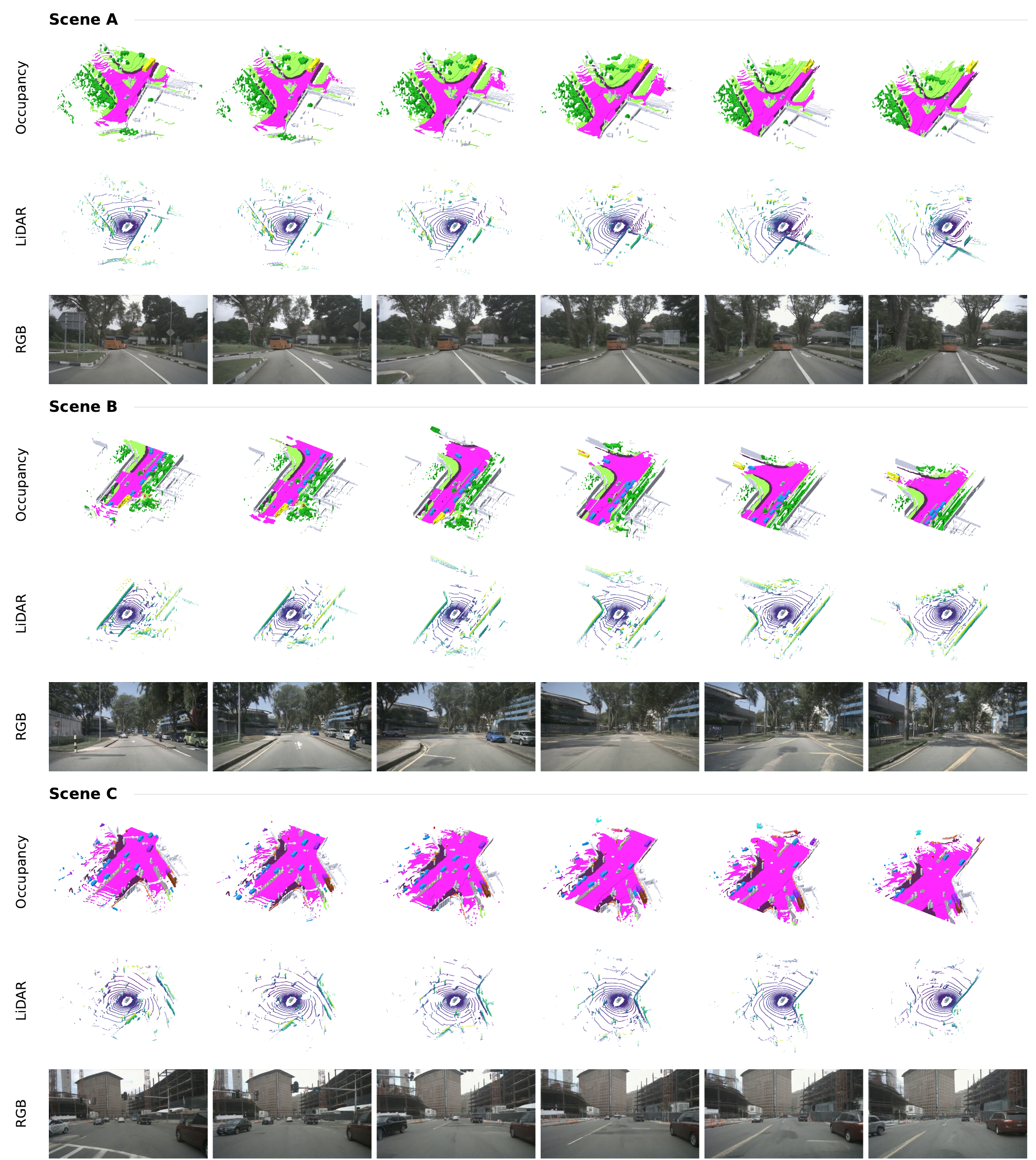}
    \caption{Occupancy, LiDAR, and RGB views for three scenes.}
    \label{fig:supp-occ-lidar}
\end{figure*}

\paragraph{Scene structure.}
Across the three scenes in Fig.~\ref{fig:supp-occ-lidar}, road corridors and roadside structures remain recognizable in both occupancy and LiDAR.
The two representations provide complementary views of these structures: occupancy depicts the spatial extent of roads and surrounding surfaces, while LiDAR depicts their sampled geometry.

\section{Scope and Limitations}
\label{sec:supp-limitations}

\begin{figure*}[p]
    \centering
    \includegraphics[width=\textwidth]{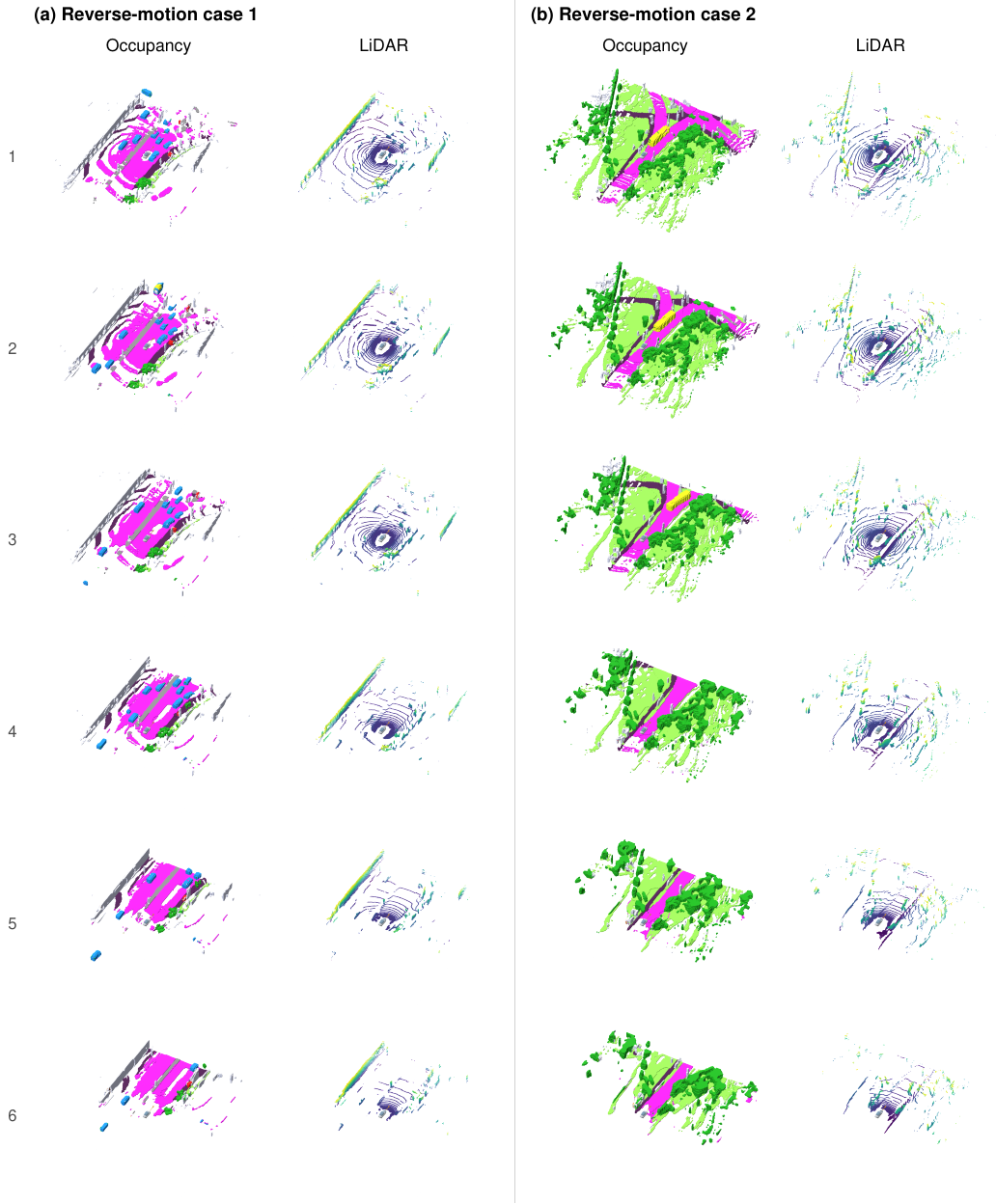}
    \caption{Two failure cases under reverse ego-motion trajectories. The cases are shown side by side, with occupancy and LiDAR predictions paired in adjacent columns. Successive frames run from top to bottom.}
    \label{fig:supp-reverse-failures}
\end{figure*}

\paragraph{Unseen motion patterns.}
Figure~\ref{fig:supp-reverse-failures} shows two failure cases under reverse ego-motion trajectories.
Across the displayed frames, surrounding occupancy becomes less complete and LiDAR geometry becomes sparser.
Our training set contains no reverse-motion trajectories, so these examples test motion patterns not seen during training.
These failures indicate limited generalization to motion patterns outside the training distribution.

% Add to the preamble:
% \usepackage{booktabs}

\begin{table}[t]
    \centering
    \label{tab:module_complexity}
    \begin{tabular}{lcc}
        \toprule
        Module & Parameters (M) & Time (ms) \\
        \midrule
        LO-VAE & 16.7 & 125.8 \\
        LO-Former & 43.2 & 102.7 \\
        \midrule
        Total & 59.9 & 228.5 \\
        \bottomrule
    \end{tabular}
    \caption{Parameter count and execution time of each module.}
    \label{tab:parameter}
\end{table}

\paragraph{Downstream task validation.}
Our downstream experiments evaluate occupancy-conditioned video generation, but do not yet establish how the generated occupancy, LiDAR, and RGB data affect perception or planning.
Systematic validation on these tasks remains future work.

% Dedicated full-page visualizations, after all text and tables.
\clearpage

\clearpage

\clearpage

\end{document}